\documentclass[10pt,twocolumn,letterpaper]{article}
\usepackage[accsupp]{axessibility}
\usepackage{times}
\usepackage{epsfig}
\usepackage{graphicx}
\usepackage{amsmath}
\usepackage{amssymb}
\usepackage[pagebackref,breaklinks,colorlinks,bookmarks=false,citecolor=cvprblue]{hyperref}
\usepackage[utf8]{inputenc}
\usepackage[T1]{fontenc}
\usepackage{url}
\usepackage{amsfonts}
\usepackage{nicefrac}
\usepackage{microtype}
\usepackage{xcolor}
\usepackage{colortbl}
\usepackage{tabularx}
\usepackage{multirow}
\usepackage{xspace}
\usepackage{booktabs}
\usepackage{makecell}
\usepackage{caption}
\usepackage[capitalize]{cleveref}
\usepackage{subcaption}

\definecolor{cvprblue}{rgb}{0.21,0.49,0.74}

\DeclareMathSymbol{@}{\mathord}{letters}{"3B}

\newcolumntype{Y}{>{\centering\arraybackslash}X}

\newcounter{magicrownumbers}
\preto\tabular{\setcounter{magicrownumbers}{0}}
\newcommand\rownumber{\stepcounter{magicrownumbers}\arabic{magicrownumbers})\,}

\newcommand{\goat}{GOAT\xspace}
\newcommand{\hmdsem}{HM3DSem\xspace}
\newcommand{\objnav}{\textsc{ObjectNav}\xspace}
\newcommand{\lnav}{\textsc{LanguageNav}\xspace}
\newcommand{\iinav}{\textsc{Instance ImageNav}\xspace}
\newcommand{\objnavfull}{ObjectGoal Navigation\xspace}
\newcommand{\ovon}{OVON\xspace}
\newcommand{\oviin}{OVIIN\xspace}

\newcommand\myquote[1]{\textit{``#1''}}

\newcommand{\moveforward}{\textsc{move\_forward}\xspace}

\newcommand{\turnleft}{\textsc{turn\_left}\xspace}
\newcommand{\turnright}{\textsc{turn\_right}\xspace}
\newcommand{\lookup}{\textsc{look\_up}\xspace}
\newcommand{\lookdown}{\textsc{look\_down}\xspace}
\newcommand{\stopp}{\textsc{stop}\xspace}

\usepackage[textsize=tiny]{todonotes}

\def\etal{\emph{et al}\onedot}

\newcommand{\xhdr}[1]{\vspace{0pt}\noindent\textbf{#1}\xspace}

\usepackage[pagenumber]{cvpr}
\usepackage[accsupp]{axessibility}  % Improves PDF readability for those with disabilities.

\def\paperID{6080}
\def\confName{CVPR}
\def\confYear{2024}

\begin{document}

\title{GOAT-Bench: A Benchmark for Multi-Modal Lifelong Navigation}

\author{
  Mukul Khanna$^{1*}$\,\,
  Ram Ramrakhya$^{1*}$\,\,
  Gunjan Chhablani$^1$\,\,
  Sriram Yenamandra$^1$\,\,
  Theophile Gervet$^2$\,\,\\
  Matthew Chang$^3$\,\, 
  Zsolt Kira$^1$\,\, 
  Devendra Singh Chaplot$^4$\,\,
  Dhruv Batra$^1$\,\,
  Roozbeh Mottaghi$^5$ \\
  % \\[0.5em]
  {\normalsize $^1$Georgia Institute of Technology \quad $^2$Carnegie Mellon University} \\ {\normalsize \quad $^3$University of Illinois Urbana-Champaign \quad $^4$Mistral AI \quad $^5$University of Washington} \\
  {\normalsize \href{https://mukulkhanna.github.io/goat-bench/}{mukulkhanna.github.io/goat-bench}} \\[-2.5em]
}

\let\svthefootnote\thefootnote
\newcommand\freefootnote[1]{%
  \let\thefootnote\relax%
  \footnotetext{#1}%
  \let\thefootnote\svthefootnote%
}

\twocolumn[{%
\renewcommand\twocolumn[1][]{#1}%
\maketitle
\centering
\captionsetup{type=figure}
\includegraphics[width=\linewidth]{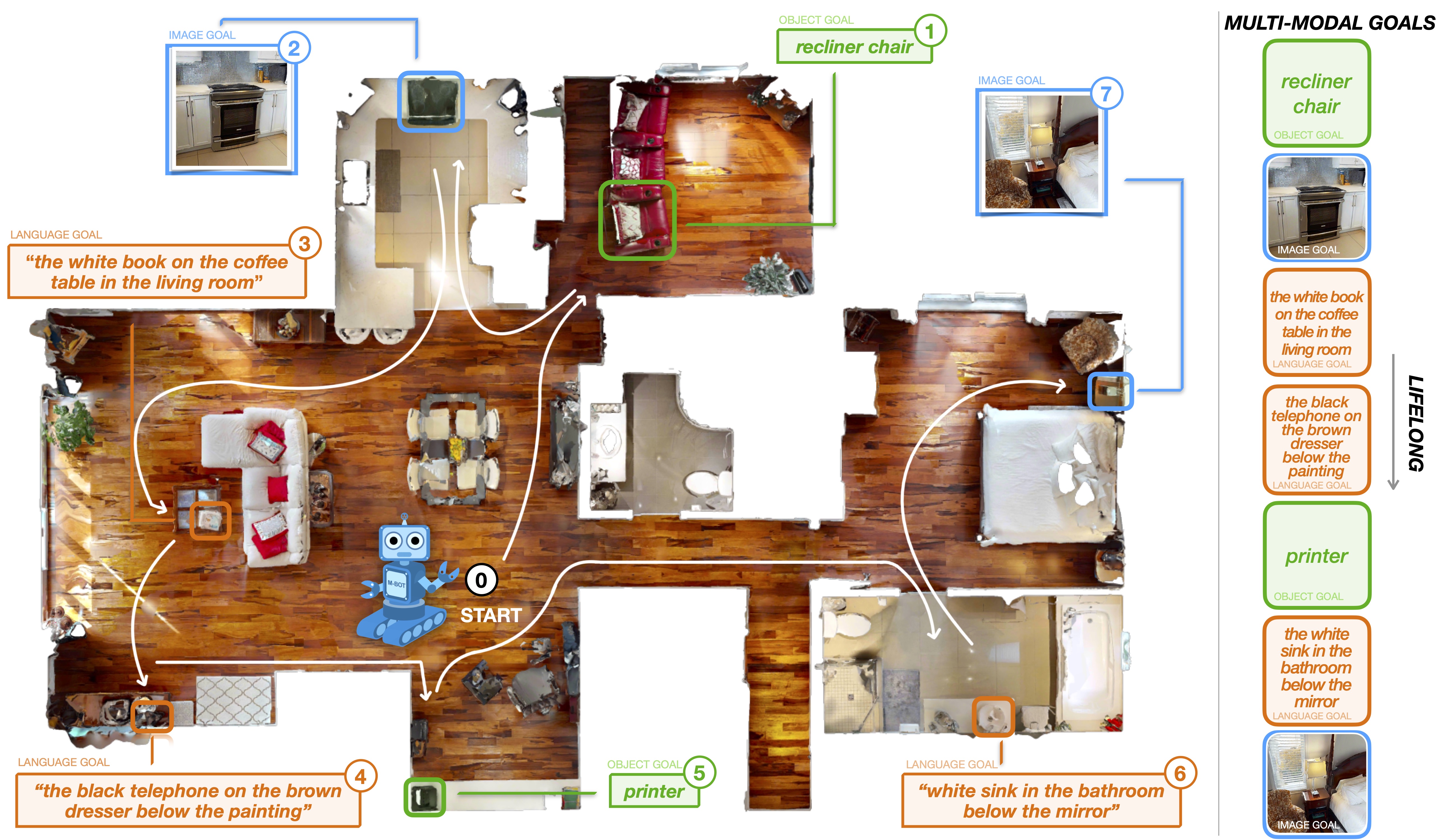}
\captionof{figure}{We study the Go to Any Thing (GOAT) task, which involves agents navigating to a sequence of open vocabulary goals specified through any of the three modalities – category name, a language description, or an image.
We propose GOAT-Bench, a benchmark for the GOAT task, where we evaluate modular and monolithic, explicit and implicit map-based navigation approaches.
In the above example, we task the agent with sequentially navigating to 1) a recliner chair (from a closed set of \textit{k} categories), 2) the oven shown in the picture, 3) \myquote{the white book on the coffee table in the living room}, and some other objects in the scene. The goal of the benchmark is to facilitate progress towards building such universal, multi-modal, lifelong agents. 
}
\label{fig:teaser}
}]

\freefootnote{$^*$Equal contribution}

\begin{abstract}

% In an era of increasing human-robot interaction, there is a pressing need for intelligent agents that can seamlessly navigate, find, and interact with objects. For these agents to serve as versatile and adaptable companions, they need to be flexible enough to understand and find objects regardless of how they are specified by humans, including supporting multi-modal (language and vision) queries. Additionally, they need to be efficient enough to utilize past experience to become better at navigating to multiple goals – one after the other. To address this need, 
% \red{We present the Go To Any Thing (GOAT) task. This task requires embodied agents to navigate to any object, irrespective of whether they are provided with a category name, a natural language description, or an image. To support the GOAT task, we introduce a comprehensive benchmark that includes Reinforcement Learning (RL) and heuristic-based baselines, including modular and end-to-end architectures, as well as memory-based and memory-less approaches. This diverse set of baselines allows for a comprehensive evaluation of the task's complexity and enables further research in this field. Additionally, to facilitate the training and evaluation of these methods, we contribute a GOAT-Bench episode dataset for HM3D simulated scenes by integrating open vocabulary variants of ObjectNav and InstanceImageNav datasets, along with a novel LanguageNav dataset.}

\noindent The Embodied AI community has made significant strides in visual navigation tasks, exploring targets from 3D coordinates, objects, language descriptions, and images. However, these navigation models often handle only a single input modality as the target. With the progress achieved so far, it is time to move towards universal navigation models capable of handling various goal types, enabling more effective user interaction with robots. To facilitate this goal, we propose GOAT-Bench, a benchmark for the universal navigation task referred to as GO to AnyThing (GOAT). In this task, the agent is directed to navigate to a sequence of targets specified by the category name, language description, or image in an open-vocabulary fashion. We benchmark monolithic RL and modular methods on the GOAT task, analyzing their performance across modalities, the role of explicit and implicit scene memories, their robustness to noise in goal specifications, and the impact of memory in lifelong scenarios.

\end{abstract}

\vspace{-21pt}
\section{Introduction}
\label{sec:intro}

%Navigation is a fundamental capability of mobile intelligent systems. 
In recent years, the Embodied AI community has established standardized evaluation metrics and benchmarks for navigation~\cite{anderson2018evaluation,batra2020objectnav,anderson2018vision,chen2020soundspaces} and developed novel algorithms and architectures~\cite{wortsman2019learning,wijmans2019dd,chaplot2020object,shah2023vint}. Notably, four different variants of navigation have emerged, depending on how the goal is specified – point-goal navigation (PointNav) % for 3D coordinates 
\cite{savva2019habitat,Ye2020AuxiliaryTS,ramakrishnan2020occant,zhao2021surprising}, object-goal navigation (ObjectNav) ~\cite{deitke2020robothor,habitatchallenge2023,ramakrishnan2022poni,gadre2023cows}, image-goal navigation (ImageNav) ~\cite{zhu2017target,krantz2023navigating,yadav2023ovrl,chaplot2020neural}, and langauge-goal navigation (referring expression or step-by-step instructions)~\cite{qi2020reverie,anderson2018vision}.

While significant progress has been made in task-specific solutions for these tasks, it is time to systematically study universal navigation methods capable of seamlessly handling goals across all of these modalities. Such a universal navigation system is crucial, as it is infeasible to specify all types of goals by a single modality.
%typically lacks the necessary expressiveness for specifying various users when specifying targets.
For instance, consider image-goal navigation ~\cite{krantz2023navigating}, where users specify the goal using an image of the target object. Capturing images of all objects within a house is infeasible if users intend to deploy the robot in a household setting. In object-goal navigation (e.g.,~\cite{deitke2020robothor}), providing the object category alone might lack the required specificity. For instance, a user might need a \textit{plate with a red pattern}, and merely providing the \textit{plate} category is insufficient to convey this level of detail.

In addition, prior works on navigation have focused on building solutions for episodic settings, \ie in each episode the agent is spawned in an indoor environment and tasked with navigating to one instance of an object category with no past memory from the environment.
However, in real-world scenarios these agents will predominantly operate in indoor environments for extended periods of time (\ie a lifelong setting), where we expect them to leverage past experiences within the same environment to become efficient over time.
%we expect them to get better over time. 
Doing so requires the ability to recall previously encountered objects and specific areas within houses, enabling them to navigate more efficiently when a new goal is specified.

Towards developing a universal, multi-modal, lifelong navigation system, we introduce a benchmark named GOAT-Bench, designed to accommodate target object specifications across multiple modalities and be capable of leveraging past experiences in the same environment \ie operate lifelong.
% (e.g., \textit{plate}), image goals
% (e.g., an image of a plate)
% (e.g., \textit{a plate next to the oven with red patterns})
\cref{fig:teaser} illustrates an example episode in GOAT-Bench. An embodied agent is spawned in a new environment and tasked with locating a recliner chair (object category goal) initially. Subsequently, it is directed to find an oven specified through an image (image goal). It is then instructed to locate \myquote{the white book on the coffee table in the living room} (language goal) and subsequently find other objects throughout the scene.
We construct GOAT-Bench using 181 HM3DSem~\cite{yadav2022habitat} scenes, 312 object categories, and $680k$ episodes. GOAT-Bench has two notable features: 

\begin{itemize}
    \item \textbf{Open vocabulary, muti-modal goals}: it is an open vocabulary benchmark, enabling the incorporation of a broad range of targets, including those not encountered during training. This is a departure from prior work, that is often limited to a small set of 6 to 21 categories \cite{deitke2020robothor, chang2017matterport3d, habitatchallenge2023, krantz2022instance}.
    \item \textbf{Lifelong}: each episode consists of 5 to 10 targets specified through distinct modalities (\ie image, object, or language goal). This contrasts with most prior navigation benchmarks where the scene is reset after a target is reached, providing a benchmark for evaluating lifelong learning.
\end{itemize}
%\zkn{What about the same target specified with multiple modalities? E.g. show it an image and say ``navigate to the dresser shown on the right in the image''. If these are not supported, will probably have to re-term it because multi-modal heavily implies that. }.
% 
%In our experiments, we evaluate two classes of methods using this new benchmark. We compare top-performing end-to-end trained approaches with modular approaches whose components are trained separately. Additionally, we evaluate methods employing implicit memory and contrast them with those incorporating explicit memory, such as maps.
We compare two classes of methods in our benchmark: a.) Sensors-to-Action using Neural Network (SenseAct-NN): Neural network policies trained using end-to-end RL (with and without implicit memory), b.) Modular Learning methods: chaining separate modules for each task component (exploration, last-mile navigation, and object detection) to solve the task (with explicit memory).
%
%Additionally, we evaluate methods employing implicit memory and contrast them with those incorporating explicit memory, such as maps.
We find SenseAct-NN methods achieve overall higher success rates ($2.9-4.6\%$ better) compared to modular methods, but achieve poor efficiency ($4.7-9.2\%$ worse) as measured by Success Weighted by Path Length (SPL).
%
% This can be attributed to end-to-end methods not being able to effectively leverage implicit map representations.
This can be attributed to the inability of SenseAct-NN methods to build/leverage implicit map representations.
In contrast, modular methods which leverage semantic maps are more effective. % \ie $4.7-9.2\%$ better on SPL.
These results highlight an area for future research - building effective memory representations for SenseAct-NN methods.

Our comprehensive analysis underscores the general importance of memory representations for improving efficiency of both SenseAct-NN and modular methods on the GOAT task.
Specifically, we find that when given access to memory, the efficiency (SPL) of both SenseAct-NN and modular methods improves for subtasks in later stages of an episode ($\sim$1.9x for SenseAct-NN and $\sim$1.5x for modular).
% We find, modular methods we tested demonstrates a decrease in both success rate and efficiency when memory is removed. 
% %
% In contrast, the end-to-end approach exhibits greater resilience to memory elimination, with a relatively smaller drop observed.
% %ods.
% %
% Furthermore, our findings emphasize the significance of memory representations across all evaluated methods, showcasing notable improvements over time, especially for later goals.
%
We also investigate how performance of these methods vary across different modalities.
We find these methods perform poorly on language and image goals, particularly when relying on CLIP~\cite{radford2021learning} features.
This suggests the inability of CLIP~\cite{radford2021learning} features in capturing crucial instance-specific and spatial features in language and image goals.
In addition, we also study how robust these methods are to noise in goals specified to the agent – by adding gaussian noise to image goals, paraphrasing language goals, and using synonyms for object goals (e.g., sofa $\rightarrow$ couch).
We find SenseAct-NN methods to be more robust to noise compared to modular methods, with a smaller drop in performance (~\cref{sec:exp_noise_analysis}).
%
% %
% SenseAct-NN methods are more robust and do not see a large drop in performance.
% %
% The modular method, however, show less robustness to noise, potentially attributed to robustness of a pre-trained object detector~\cite{zhou2022detecting} for object goals, heuristics used for keypoint matching of image goals and expressiveness of CLIP embedding for instance-specific language goals.
%
% Such detectors typically struggle with objects specified by uncommon names, leading to reduced adaptability in the input variations.

% Can we comment this
\noindent To summarize, our contributions are:
\begin{itemize}
    \item A novel reproducible benchmark for building and evaluating multi-modal lifelong navigation systems.
    \item Benchmarking of modular and end-to-end trained methods with and without memory representations.
    \item A comprehensive analysis of these methods on memory dependency, performance across modalities, and robustness to noise.
\end{itemize}

% \noindent This benchmark makes a step toward more comprehensive and realistic navigation systems, incorporating multiple modalities, and we hope that it opens up new avenues in Embodied AI research.

\section{Related work}
\label{sec:related}

\begin{figure*}
    \centering
    \includegraphics[width=1\linewidth]{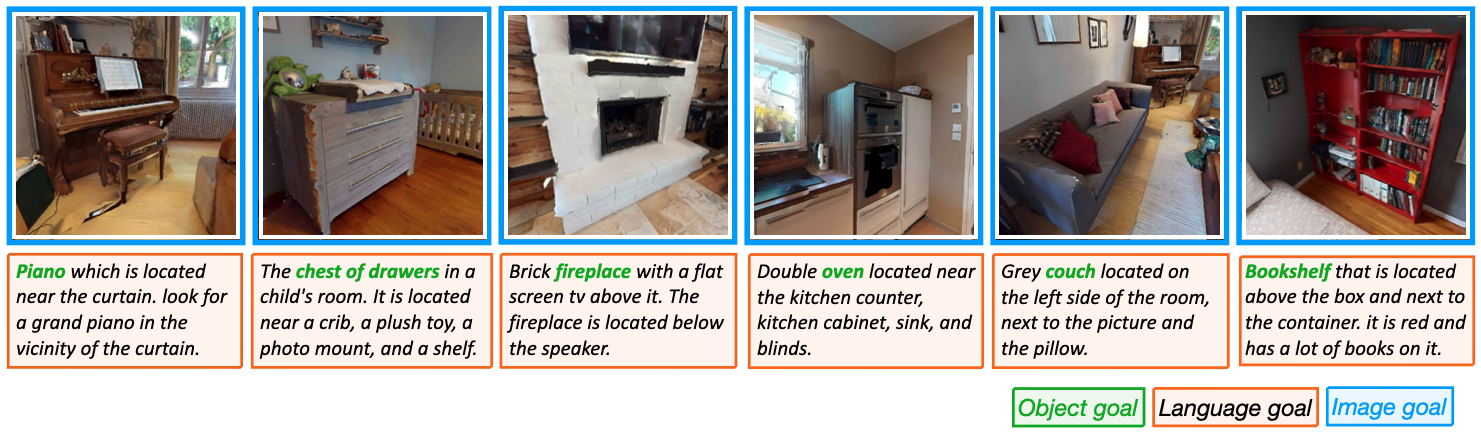}
    \caption{\textbf{Preview of the GOAT-Bench dataset.} We show multi-modal examples of goal instances from the dataset: images of objects (blue), language descriptions (orange) and object category annotations (green).}
    \label{fig:dataset_preview}
    \vspace{-1.6em}
\end{figure*}

\textbf{Navigation in virtual environments.} 
% In recent years, there's been a lot of work in the field of navigation, exploring various aspects like image~\cite{zhu2017visual}, object~\cite{objectnav_tech_report,deitke2020robothor}, language~\cite{mattersim}, and 3D coordinates~\cite{anderson_arxiv18} as navigation targets. However, existing benchmarks mostly assess navigation agent performance on each type of navigation independently. 
In recent years, most benchmarks have assessed navigation performance on individual goal types (image, object, language, and 3D coordinates)~\cite{chaplot2020neural,chaplot2020object,hahn2021no,Ramakrishnan_2022_CVPR,ramrakhya2022,wasserman2022last,wijmans2019dd,yadav_ovrl,ye2021auxiliary,al2022zero}.
Methods on these benchmarks often use modality-specific goal encoders or recognition modules. For example, one-hot encodings or detections for object goals, SuperGLUE-based keypoint matching \cite{sarlin20superglue, krantz2023navigating} or cross-view consistent encoders~\cite{croco_v2} for image goals, or linear projection for 3D coordinates~\cite{wijmans2019dd}.
These approaches are tailored to individual modalities and are unable to generalize across modalities out-of-the-box. 
In contrast, our focus is to study the performance of general-purpose architectures that can handle multiple modalities.
Recent efforts~\cite{majumdar2022zson} have tackled this problem by leveraging advances in vision-and-language aligned models (CLIP~\cite{radford2021learning}) to bridge this gap by using a single goal encoder for handling image and object goals.
However, \cite{majumdar2022zson} does not focus on longer natural language descriptions that are required to disambiguate and identify specific object instances. 
% \red{And as we show in our experiments, using CLIP goal encoding doesn't perform well when navigating to specific object instances.}
Additionally, as we show in our experiments (\cref{sec:main_results}), CLIP goal encodings don't help when navigating to specific object instances.
%Also, it only focuses on single-goal episodic settings and does not study this lifelong aspect of memory and navigation.}
%%
% Our work introduces a unified benchmark to evaluate methods that can operate with goals across multiple modalities, aiming to advance the development of general navigation agents.

\noindent \textbf{Embodied Multi-Modal Benchmarks.}
% A wide range of embodied tasks require different task specification formats, like instruction-following to accomplish long-horizon tasks that require specific state changes~\cite{mattersim,ALFRED20}, question answering~\cite{embodiedqa}, multi-modal prompts specifying different object arrangements for robot manipulation~\cite{jiang2023vima}.
% %
% ~\cite{mattersim} require embodied agents to ground language instructions in the environment to navigate to a goal by following step by step instructions.
% %
% ~\cite{ALFRED20} requires agents to follow instructions to accomplish long compositional tasks with irreversible state changes.
% %
% ~\cite{embodiedqa} requires agents to navigate and interact with the environment to answer questions about the environment.
% %
% In this work, we are interested in building embodied agents that can understand multi-modal goal specifications to navigate to specified location/object in the environment.
% %
% Work most similar to ours is ~\cite{jiang2023vima}, which proposes VIMA-Bench a benchmark for multi-modal robot manipulation tasks. 
%Embodied agents across diverse domains necessitate leveraging multiple modalities (language and vision) as input to accomplish tasks.
Existing embodied tasks~\cite{ALFRED20,embodiedqa,mattersim,jiang2023vima} require embodied agents to work with inputs from multiple modalities (language, vision, audio, etc) but they seldom have agents leveraging past experiences from the same environment, \ie through lifelong agent scenarios.
For example, ALFRED~\cite{ALFRED20} involves following instructions to achieve long-horizon tasks and EmbodiedQA~\cite{embodiedqa} requires an agent to answer a question by exploring or interacting with the environment. Both of these tasks require agents to leverage multi-modal inputs (language and image) but they are studied in single episode settings.
In contrast, our primary focus is on navigation agents capable of understanding multi-modal open-vocabulary goals in lifelong scenarios.
Most similar to our work is~\cite{wani2020multion}, where an agent is tasked to navigate to multiple objects from a closed-set of object categories in the same environment but a key difference in our work is that goals in the \goat task are multi-modal (object category, image, and language description).
%
% Meanwhile, \cite{embodiedqa} requires agents to navigate around to respond accurately to queries about the scene.
% %
% However, these tasks have unimodal goal specifications (language in this case) and are not studied in lifelong setting.
%
%
% A work closely aligning with our interests is \cite{jiang2023vima}, which introduces VIMA-Bench, a benchmark specifically designed for multi-modal robot manipulation tasks. This benchmark provides a foundational platform for evaluating and advancing the capabilities of embodied agents in multi-modal interaction and navigation

\noindent \textbf{Concurrent Work}.
In tandem with our efforts, there is concurrent work that proposes the HM3D Open-Vocabulary ObjectNav task \cite{ovon}. In contrast to object category-based single-goal-per-episode setup in \cite{ovon}, we focus on navigating to a sequence of goals specified across three different modalities.
Similarly, there is concurrent work that proposes a modular system for solving the GOAT task in real world houses for a closed set of 15 object categories \cite{chang2023goat}. In contrast to \cite{chang2023goat}, our work focuses on a practical, open-vocabulary setting and contributes a reproducible benchmark that the community can use to facilitate progress towards universal navigation agents.
Having access to a reproducible benchmark in simulation allows us to ask and answer questions 
% related of \goat task,
about various aspects of these universal navigation agents,
such as the role of effective memory representations, compare against existing and future methods, and analyze robustness of these methods to noise across modalities.
Such questions remain unanswered in~\cite{chang2023goat} due to the time-intensive nature of real world evaluations.
%
% Furthermore, efforts towards building such a reproducible benchmark is complementary to~\cite{chang2023goat} and should be done in addition to real-world benchmarking not in replacement.
Furthermore, the efforts to construct such a reproducible benchmark aligns with the objectives of \cite{chang2023goat} and should be viewed as a complementary effort, meant to augment, not replace, real-world benchmarking.
\vspace{-0.3em}

%
% Additionally, while \cite{chang2023goat} focuses on real-world evaluation of only} modular approaches with explicit semantic maps, we also evaluate the potential of \red{two types of sensor-to-actions neural network policies trained using RL on this} task. 

\section{Task}
\label{sec:task}

%In this section, we describe the Go to Any Thing (GOAT) task.
In the Go to Any Thing (GOAT) task, an agent is spawned randomly in an unseen indoor environment and tasked with sequentially navigating to a variable number (in 5-10) of goal objects, described via the category name of the object (\eg `couch'), a language description (\eg \myquote{a black leather couch next to coffee table}), or an image of the object uniquely identifying the goal instance in the environment. We refer to finding each goal in a GOAT episode as a \textit{subtask}. Each GOAT episode comprises 5 to 10 subtasks.

We set up the \goat task in an open-vocabulary setting; unlike many prior works, we are not restricted to navigating to a predetermined, closed set of object categories \cite{deitke2020robothor, habitat_challenge2022,krantz2022instance,wani2020multion,objectnav_tech_report}. The agent is expected to reach the goal object $g^k$ for the $k^{th}$ subtask as efficiently as possible within an allocated time budget. Once the agent completes the $k^{th}$ subtask by reaching the goal object or exhausts the allocated time budget, the agent receives next goal $g^{k+1}$ to navigate to. This contrasts with most prior navigation benchmarks~\cite{deitke2020robothor,objectnav_tech_report,krantz2022instance,anderson_arxiv18} where the scene/episode is reset once the agent reaches the goal. Chaining multi-modal navigation goals enables us to benchmark lifelong learning methods that leverage past agent experience in the same environment.

We use HelloRobot's Stretch robot embodiment for the \goat agent. The agent has a height of 1.41m and base radius of 17cm.
At each timestep, the agent has access to an 360 x 640 resolution RGB image $I_t$, depth image $D_t$, relative pose sensor with GPS+Compass information $P_t = (\delta x, \delta y, \delta z)$ from onboard sensors, as well as the current subtask goal $g^{k}_t$, $k$  $\forall$ $\{1, 2,...,5-10\}$.
% This goal could be the name of an object category (referring to all instances of the category), a sentence describing a specific object instance, or an image of that instance.
The agent's action space comprises \moveforward (by 0.25m), \turnleft and \turnright (by 30º), \lookup and \lookdown (by 30º), and \stopp actions. A sub-task in a GOAT episode is deemed successful when the agent calls \stopp action within $1$m euclidean distance from the current goal object instance – within a budget of 500 agent actions (per sub task). %Note that at each time step, the agent only has access to the goal of current subtask. %, and future goals are not visible till the end of the current subtask.  %This allows us to study sequential multi-modal navigation. %In future, this benchmark could be used to study multi-goal planning by trivially changing the visibility of the goals to the agent. 

% Currently, most embodied navigation tasks only focus on individual object category or image goal specifications\zkn{Seems out of place in the flow of things. This has been mentioned before. }. Note that, through this work, we also propose a novel intermediate LanguageNav task where the agent is tasked with finding object instances described by language\zkn{This is confusing. What do you mean by ``intermediate'' task? What is the purpose of this task, compared to what you've described above? You've already mentioned that natural language is an input; is this what you mean? If so, I would put in above narrative when you first introduce that concept. }. LanguageNav follows a similar setup as aforementioned \goat task in terms of sensor specifications, action space, agent budget, and success criteria.

% \zkn{Currently there are a lot of things going on. Multi-modal, lifelong, memory, intermediate LanguageNav task, etc. I'm not clear on the coherent story. There has to be a natural flow set up in the intro that introduces these, and then the Task section can mirror the structure with details. }

\section{Dataset}
\label{sec:dataset}
\begin{table}
    \centering
    \resizebox{1\linewidth}{!}{
        \setlength\tabcolsep{2pt}
        \begin{tabular}{@{}llccccccccccc@{}}
            \toprule
            & & \multicolumn{2}{c}{\textsc{Train}} & &
            \multicolumn{2}{c}{\textsc{Val Seen}} & & 
            \multicolumn{2}{c}{\textsc{Val Seen Synonyms}} & & \multicolumn{2}{c}{\textsc{Val Unseen}} \\
            \cmidrule{3-4} \cmidrule{6-7} \cmidrule{9-10} \cmidrule{12-13}
            & Dataset & Categories & Goals
                & & Categories  & Goals & & Categories  & Goals & & Categories & Goals  \\
            \midrule
            \\[-10pt]
            %& \rownumber & Skill Chain & $13.20$ & $-$ & & $16.80$ & $-$ & & $12.65$ & $-$ \\
            & RoboTHOR Challenge \cite{deitke2020robothor}
             & $12$ & $420$ &
             & $12$ & $105$ &
             & $-$ & $-$ &
             & $-$ & $-$ \\
            & ObjectNav-MP3D \cite{chang2017matterport3d} 
             & $21$ & $7509$ &
             & $21$ & $1316$ &
             & $-$ & $-$ &
             & $-$ & $-$ \\
            & ObjectNav-HM3D \cite{habitatchallenge2023}
             & $6$ & $5216$ &
             & $6$ & $1168$ &
             & $-$ & $-$ &
             & $-$ & $-$ \\
            & InstanceImageNav-HM3D \cite{krantz2022instance}
             & $6$ & $3516$ &
             & $6$ & $780$ & 
             & $-$ & $-$ &
             & $-$ & $-$ \\
             \midrule
            & OVON \cite{ovon}
             & $280$ & $10987$ &
             & $79$ & $2219$ &
             & $50$ & $1177$ &
             & $49$ & $1278$ \\
            % & \rownumber & OVIIN
            %  & $264$ & $7724$ &
            %  & $70$ & $1256$ &
            %  & $48$ & $736$ &
            %  & $46$ & $924$ \\
            % & \rownumber & LanguageNav
            %  & $229$ & $5438$ &
            %  & $62$ & $887$ &
            %  & $40$ & $634$ &
            %  & $38$ & $470$ \\
             \midrule
            & GOAT-Bench
             & $193$ & $13025$ &
             & $52$ & $1760$ & 
             & $31$ & $877$ &
             & $36$ & $1282$ \\
            \bottomrule
            \end{tabular}
    }
    \vspace{5pt}
    \caption{\textbf{Dataset statistics for popular embodied navigation benchmarks.} 
    % We present numbers across three evaluation splits: \textsc{Val Seen} (object categories seen during training), \textsc{Val Seen Synonyms} (object categories synonymous to those seen during training), and \textsc{Val Unseen} (object categories not seen during training). 
    GOAT-Bench has at least ${\sim}9$x more object categories for training (21 vs 193) and about ${\sim}6$x more for validation (21 vs 119) than prior closed-set navigation benchmarks. } 
    \label{tab:dataset_stats}
    \vspace{-1.8em}
\end{table}

In this section, we describe the procedure used to build an open-vocabulary GOAT-Bench.
We use real-world 3D scans from HM3DSem~\cite{yadav2022habitat}, consisting of $145$ training and $36$ validation scenes. 
%
% To construct this dataset we combine open-vocabulary object, instance image, and language goals. 
In total, GOAT-Bench consists of $264$ training categories and a total of ${\sim}13k$ goal specifications across a total of $725k$ training episodes (with $5k$ episodes per training scene). This is in contrast with most prior embodied navigation datasets that focus on a closed set of 6 to 21 object categories – with goals for one modality. We compare the scale of our dataset against prior work in \cref{tab:dataset_stats}. GOAT-Bench has about $9$x more object categories for training and about $6$x more for validation than prior closed-set datasets.
%\red{Because GOAT-Bench has annotations for 3 modalities per instance, it is significantly bigger than prior works in terms of instance and annotation diversity.}
%
Next, we describe how we generate goals for each modality.

\noindent \textbf{Open-Vocabulary ObjectNav goals (OVON)}. The OVON task from HM3D-OVON~\cite{ovon} has embodied agents navigate to object goals from an open vocabulary (as opposed to from a fixed set). This involved extending the \objnavfull task from~\cite{yadav2022habitat,objectnav_tech_report} to an open-vocabulary setting with hundreds of categories by leveraging the dense semantic annotations provided in HM3DSem~\cite{yadav2022habitat}.
Specifically, HM3D-OVON\cite{ovon} extends the 6 category \objnav dataset from~\cite{yadav2022habitat} to $280$ object categories for training and $179$ object categories for evaluation (similar to~\cite{ovon}).
To do so, they leverage the ground truth semantic annotations from HM3DSem dataset~\cite{yadav2022habitat} and sample objects which are large enough to be visible, \ie objects with frame coverage $\geq 5\%$ from any viewpoint within $1m$ of the object. Frame coverage refers to the ratio of goal object’s pixels to the total number of pixels.
%
% Using this, we generate total $280$ unique object goals for training and $179$ object goals in the evaluation split 
We use these goals from ~\cite{ovon} – including both seen and unseen categories – to test generalization to novel objects (see supplementary for full list).

\noindent \textbf{Open-Vocabulary Instance-ImageNav goals (OVIIN)}. \iinav \cite{krantz2022instance} is the task of navigating to object instances specified by images for the canonical $6$ \objnav categories in HM3D scenes~\cite{ramakrishnan2021hm3d,yadav2022habitat}.
However, in this work, we are interested in studying generalization to a wide variety of novel, unseen instances and categories of image goals.
We do so by extending the \iinav task to an open-vocabulary setting by creating training and evaluation splits using same heuristics as OVON goals to build the \oviin goals.
As a result, we generate a total of $7.7k$ image goal instances across $264$ training categories and $2.9k$ instances across $164$ validation categories for evaluation. 
We show some samples from the OVIIN dataset in \cref{fig:dataset_preview}. Refer~\cref{sec:iinav_dataset} for details on goal image sampling. 
%
%\ramtodo{Numbers here do not match the numbers in the table. Also, I think we should mention why/how some categories are dropped in going from OVON -> OVIIN due to visibility issues.}

\begin{figure}
    \centering
    \includegraphics[width=1\linewidth]{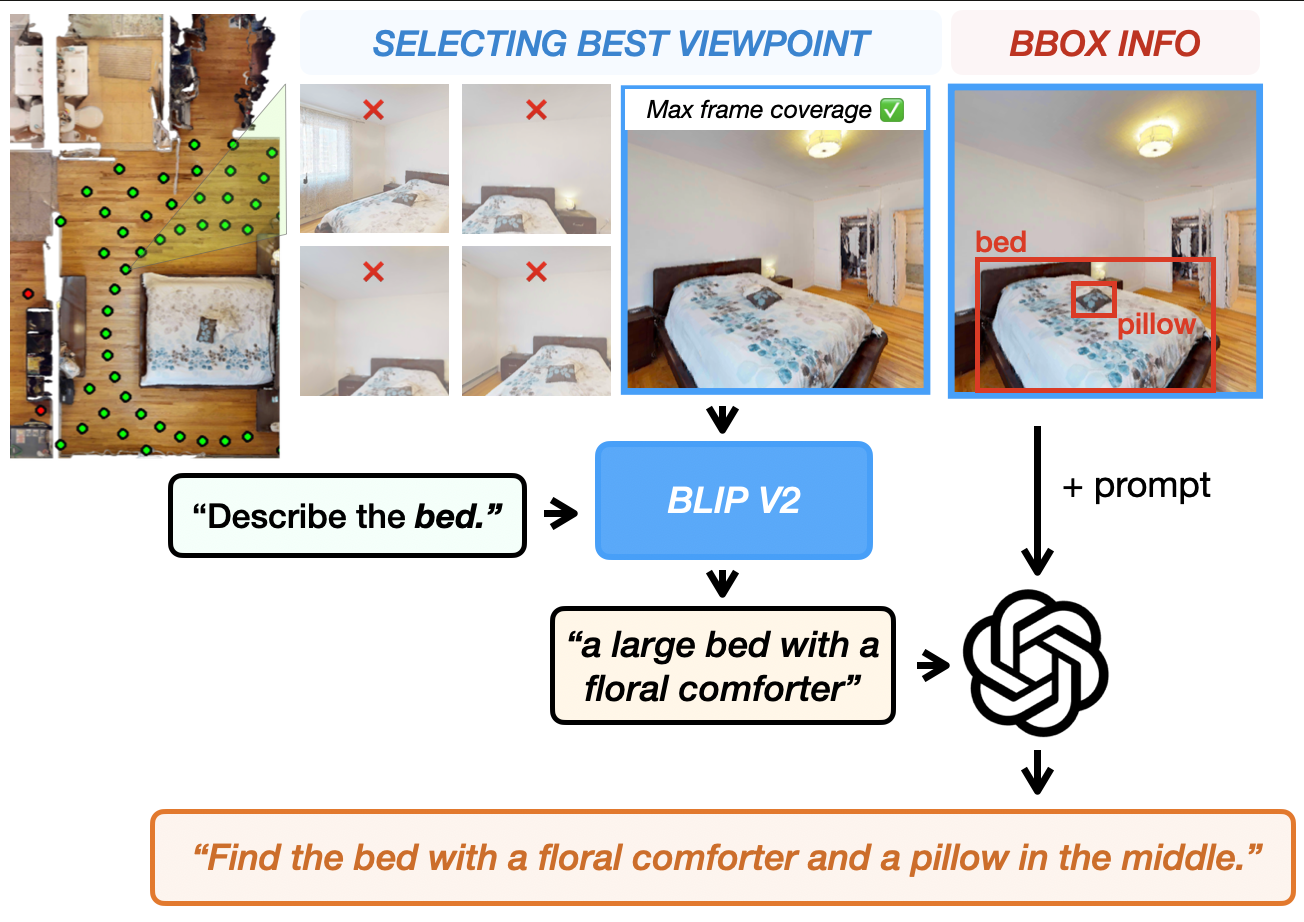}
    \caption{\textbf{LanguageNav dataset generation pipeline.} We automatically generate language descriptions for object goals by leveraging VLMs, LLMs and ground truth information from simulator. We first capture an image of the goal object from a valid viewpoint. Next, we retrieve spatial and semantic information of the nearby objects from the simulator. We then prompt BLIP-2~\cite{li2023blip2} to extract appearance attributes of the object. These are then combined to prompt ChatGPT-3.5 to output a language description of the goal.}
    \label{fig:languagenav_generation}
    \vspace{-1.6em}
\end{figure}

\noindent \textbf{Language goals (LanguageNav)}.
LanguageNav involves agents navigating to objects described by natural language (\eg \myquote{plate with a red flower pattern}).
Prior works in language-based navigation either provide verbose step-by-step instructions to reach a goal ~\cite{anderson2018vision,ku2020room} or limited human annotations for evaluations~\cite{qi2020reverie,ALFRED20,ku2020room} to describe the goals.
%
% In most common household settings, providing the object category alone is oftentimes not sufficient to convey the specific instance to a navigation agent. To use the same example, if a user needs `a plate with red flower pattern', merely providing \emph{plate} as the goal object category will not be sufficient to convey this level of detail.
%
However, collecting human annotations for language descriptions for thousands of object instances and scaling them with increasing scene dataset size is challenging and expensive.
Generating descriptions for these objects in an automated fashion, on the other hand, is also challenging.
It requires distilling object's visual appearance, spatial, and semantic context into coherent sentences.
This includes information about the object, its attributes (color, shape, material properties, etc.), its spatial relationship to surrounding objects, what room it is in, etc.
To tackle these challenges, we present an automatic pipeline for generating language descriptions by leveraging ground truth semantic and spatial information from simulators along with reasoning capabilities of popular vision-and-language (VLM) and large language models (LLM).
As shown in \cref{fig:languagenav_generation}, for each object goal instance in \ovon, we sample the viewpoint image of a goal with the maximum frame coverage (\ie ratio of goal object's pixels to total number of pixels), extract semantic and spatial information from the simulator \cite{savva2019habitat} such as names and 2D bounding box coordinates of visible objects.
For each sampled goal instance, we prompt BLIP-2~\cite{li2023blip2} to extract attributes like color, shape, material, etc.
Finally, we combine the spatial and semantic information from the simulator with object attribute metadata from BLIP-2~\cite{li2023blip2} predictions and use that to prompt ChatGPT-3.5 to output a language description of the object instance.
%\zkn{Since you're using it for evaluation (held-out), how do you know if it's accurate enough that the accuracy results are meaningful? Did you sample it somehow and validate?} .
%
Using this pipeline, we generate a total of $5.4k$ unique language goal instances from $225$ training categories and $1.9k$ goal instances from $137$ held out categories. We show examples for these in \cref{fig:dataset_preview} and ~\cref{sec:lnav_dataset}. 

\noindent \textbf{GOAT-Bench Dataset Episode Generation}.
Combining the above mentioned open vocabulary datasets provides us with (category name, language description, and image) tuples associated with each goal object instance. These are used to generate training and evaluation episodes with multi-modal goal specifications for lifelong navigation. 
%
%Our episode generation process is similar to the \objnav task with changes to support multimodal goal specifications and lifelong navigation.
%
Each episode consists of a scene, the agent's starting position (at timestep $t = 0$), and sequences of 5 to 10 (sub-task) goals across three modalities.
To generate each episode, we first uniformly sample a number of subtasks between 5 to 10. For each subtask, we uniformly sample a goal modality (category, description, or image), and then randomly sample a goal instance – uniformly across all categories.
We then randomly sample a starting position that satisfies the following constraints: a.) all subtask goal locations are on the same floor as the starting position as we do not expect the agent to climb stairs, and b.) distance to nearest goal location for first subtask must lie between $1m$ to $30m$. 
This is similar to the episode generation process for the \objnav task \cite{habitat_challenge2022}.
%We ensure that all goals of an episode are on the same floor of the environment as we do not expect the agent to climb stairs in indoor environments.
%
See~\cref{fig:teaser} for an example of what goals for a single episode look like. 
Following this procedure, we generate $5k$ train GOAT episodes ($25$ to $50k$ subtasks) per scene for 145 training scenes. For the validation set, we generate 10 GOAT episodes ($50$ to $100$ subtasks) per scene for 36 val scenes.

\noindent \textbf{Evaluation Splits}.
To test generalization of navigation agents we evaluate these agents in unseen environments, which means each goal instance is novel.
In addition, to test generalization to objects at different levels, we generate $3$ evaluation splits: \textsc{Val Seen}, \textsc{Val Seen Synonyms}, and \textsc{Val Unseen} by manually segregating object categories depending on whether they were observed during training.
\begin{itemize}
    \item \textsc{Val Seen} - goals generated using object categories seen during training. 
    \item \textsc{Val Seen Synonyms} - goal categories synonymous to those seen during training (\ie ``couch'' category seen during training, evaluated on ``sofa'' during evaluation).
    \item \textsc{Val Unseen} - goals generated using object categories not seen during training.
\end{itemize}

% \red{Note that all of these splits consist of goal object instances that were not seen during training – because these come from 3D scans of houses that do not share the exact set of objects.}

\section{Baselines}

In this section, we present multi-modal policies trained on the \goat task using the  \hmdsem~\cite{yadav2022habitat} scene dataset in the Habitat simulator~\cite{savva2019habitat}.
We benchmark two types of methods: 1) Modular methods: semantic mapping and planning-based, and 2) Reinforcement Learning: sensor-to-action using neural network (SenseAct-NN) policies trained using RL.

\subsection{Modular Baseline}

Modular navigation approaches have emerged as a popular paradigm for training policies for various Embodied AI tasks~\cite{chaplot2020learning, chaplot2020neural, chaplot2020semantic, chaplot2020object, Krantz_2021_ICCV, chaplot2021seal, georgakis2021learning, hahn2021nrns, min2021film, sarch2022tidee, ramakrishnan2022poni}.
The key idea in these approaches is to decouple low-level control for navigation from goal recognition. This allows us to have separate modules dedicated for each task component (like detection, exploration, last-mile navigation), which are then chained to solve the task.
Prior work~\cite{chaplot2020semantic} builds a top-down semantic map by projecting first-person semantic predictions with depth. It then selects an exploration goal on the semantic map using the goal query through a learned or heuristic exploration policy, and plans low-level actions to the goal.

\noindent \textbf{Modular \goat} \cite{chang2023goat} extends prior modular navigation methods~\cite{chaplot2020neural,chaplot2020learning,chaplot2020object} to handle multi-modal goal prompts (\ie object category, language, and image goals).
Specifically, they build an instance-specific memory (alongside the semantic memory) by clustering together projected pixels of same categories on the semantic (top-down) map \cite{chang2023goat}. This instance memory captures egocentric views and CLIP features of object instances seen during exploration. Depending on the current goal modality, the agent then matches the current goal (image or CLIP embedding of description) with object instances – through keypoint-matching for image-to-image matching and cosine similarity for language-to-image feature matching. Instances with the best matching score are then localized and marked as goals for the agent to navigate to. Note that this method assumes access to the object category information to filter out instances during goal matching.

\noindent \textbf{Modular CLIP on Wheels (CoW)}. Similar to \cite{chang2023goat}, we also present results with CoW \cite{gadre2022cows}, that uses only CLIP features to match (image, object, and language) goals against all of the images seen during exploration to localize the goal. 

\subsection{SenseAct-NN Baselines}
\label{sec:e2e_baseline}

In addition to evaluating modular approaches, we also train sensor-to-action neural network policies using RL for the \goat task.
Specifically, we consider two methods:

\noindent \textbf{SenseAct-NN Skill Chain}. Learning a single sensor-to-action neural network (\ie monolithic policies) using end-to-end RL for \goat task is difficult due to the long horizon nature of the task.
As an alternative, we train \emph{individual} navigation policies for each \goat subtask – \objnav, \iinav, and \lnav. We combine these using a high level planner which executes one of the available policies based on the navigation goal modality at each timestep.
Specifically, we extend the policy architecture from~\cite{khandelwal2021simple} and use a simple CNN+RNN policy architecture.
To encode RGB input $(i_t =$ CNN$(I_t))$, we use a frozen CLIP~\cite{radford2021learning} ResNet50~\cite{he_cvpr16} encoder.
The GPS+Compass inputs, $P_t = (\Delta x, \Delta y, \Delta z)$, and $R_t = (\Delta \theta)$,
% between successive steps relative to start of the episode $(x_0, y_0, z_0, \theta_0)$.
%
are passed through fully-connected layers $p_t =$ FC$(P_t), r_t =$ FC$(R_t)$ to embed them to 32-d vectors.
Finally, we convert the goal observation to $d$-dimensional vector using a modality-specific goal encoder $g^{(m)}_t =$ ENC$(G^{(m)}_t)$.
All of these input features are concatenated to form an observation embedding, and fed into a 2-layer, 512-d GRU at every timestep to predict a distribution over actions $a_t$ - formally, given current observations $o_t = [i_t, p_t, r_t, g_t]$, $(h_t, a_t) =$ GRU$(o_t, h_{t-1}) $.

\noindent For each subtask type in \goat task we ablate the choice of visual encoder and goal encoder for training task-specific policies and choose the one which performs the best for that subtask (refer~\cref{sec:main_results} for results).
For \objnav goals, we use a frozen CLIP~\cite{radford2021learning} text goal encoder and CLIP ResNet50~\cite{he_cvpr16} visual encoder, and for \lnav, we use a BERT~\cite{bert} sentence goal encoder and CLIP ResNet50~\cite{he_cvpr16} visual encoder. For \iinav, we use the recently released CroCo-v2~\cite{croco_v2} to generate cross-view consistent goal and visual embeddings.
We train each of these policies using VER~\cite{wijmans2022ver} till convergence on task-specific datasets (refer ~\cref{sec:skill_chain_hyperparam} for training details).

\noindent \textbf{SenseAct-NN Monolithic Policy}. We also benchmark a monolithic sensor-to-action neural network policy trained using RL for the \goat task.
Training an effective multi-modal policy capable of leveraging past experience from previous \goat subtasks requires two important properties: a.) a multimodal goal encoder which can map goals from different modalities into a common latent space for the policy (e.g. CLIP), and b.) an implicit or explicit memory representation for capturing past experience.
For encoding the goals, we use CLIP~\cite{radford2021learning} text and image encoders. Because CLIP is trained with a vision-and-language alignment loss, we expect it to output meaningful representations in a common latent space for effective goal encoding.
Next, to leverage past experience we carry forward hidden state of the policy from last subtask $h^{(s_{t-1})}_{T}$ as initial hidden state for a new subtask $h^{(s_t)}_{0}$ in a single \goat episode.
Wijmans~\etal \cite{wijmans2023emergence} showed blind agents modeled using RNNs are capable of building map-like representations for the PointNav~\cite{anderson_arxiv18} task. 
Motivated by these experiments, we expect maintaining an RNN hidden state across subtask provides our policy with a implicit memory representation which can be effectively used for efficient navigation.
%Skill chaining baseline does not leverage the memory across \goat subtasks which makes it inefficient.
%
Towards this end, we extend the policy from~\cite{khandelwal2021simple} to train a monolithic policy with CLIP as our goal encoder $g^{(m)}_t =$ ENC$(G^{(m)}_t)$ and maintain hidden states across subtasks during training. 
We train this policy using VER~\cite{wijmans2022ver} for $500$ million steps on \goat train dataset; refer~\cref{sec:end_to_end_hyper} for more details.

\section{Results}

% For our experiments, we use HelloRobot's Stretch robot embodiment. The agent has a height of 1.41m, base radius of 17cm, and a 360 x 640 resolution RGB-Depth camera at a height of 1.31m.
%
For our experiments, we report two metrics – success rate (SR) and Success Weighted by Path Length (SPL). Success rate represents the percentage of sub-tasks where the agent successfully navigates to a goal. Efficiency, on the other hand is measured using SPL \cite{batra2020objectnav} – where the shortest path for each sub-task is considered from the final location of the agent from the previous sub-task to the goal location for the current sub-task. For the first sub-task, this corresponds to the starting position of the episode.

% \subsection{Experimental Setup}

% For our experiments, we use HelloRobot's Stretch robot embodiment. The agent has a height of 1.41m, base radius of 17cm, and a 360 x 640 resolution RGB-Depth camera at a height of 1.31m.

% We evaluate these agents on two primary metrics – success rate (SR) and efficiency – through Success Weighted by Path Length (SPL). We report the SR and SPL as an average of success rates and SPLs across all sub-tasks. For computing the SPL, we use the position of the agent where it calls STOP for the previous sub-task as the starting position of the current sub-task. For the first sub-task, this corresponds to the starting position of the episode.

\subsection{Modular vs. SenseAct-NN approaches}
\label{sec:main_results}

We present comparisons between modular approaches (with explicit maps) and SenseAct-NN RL approaches (with and without implicit maps) in \cref{tab:mod_vs_end2end_results}. In terms of success rate, we observe that the SenseAct-NN Skill Chain baseline (row 3) outperforms all other baselines (that do not use ground truth semantics, shown in row 1) across all three validation splits. It also appears to be generalizing better to unseen instances and categories – performing better than the modular baselines (row 2) by at least an average margin of about $4\%$ across all splits. However, this baseline does not do as well on SPL – on average $6.6\%$ lower than the best modular (GOAT) baseline. This is because it does not maintain any memory across sub-tasks to keep track of previously encountered objects and regions of the scene. Specifically, as we have separate navigation policies for each modality, the policy hidden state is not propagated across sub-tasks. 

On the other hand, the Modular GOAT~\cite{chang2023goat} (row 2), which maintains an explicit semantic and instance map of the environment, does much better on SPL ($6.6\%$ than SenseAct-NN Skill Chain and $10.9\%$ better than SenseAct-NN Monolithic). After the agent has sufficiently explored the scene, it is able to leverage this memory for localizing new goal instances in already seen parts of the map and navigating to them directly. 
Modular GOAT also does better than the modular CLIP on Wheels (CoW) baseline (row 3) – highlighting the usefulness of maintaining an instance-specific memory, using category information to filter instances, and relying on image keypoints instead of CLIP features for matching against image goals.
To decouple limitations of Modular GOAT's perception module (for object detection and map building) from the instance-to-goal matching, heuristic planning, and last-mile navigation, we also show its results with ground truth semantics (row 1). This reflects an average improvement of ${\sim}30\%$ in success and ${\sim}22.0\%$ in SPL. 

We also observe that the SenseAct-NN Monolithic policy (row 4) does not perform well compared to the other baselines. 
We hypothesize this is due to: 1.) CLIP's limited efficacy in capturing instance-specific features for language and image goals, 2.) difficulty of learning effective long horizon navigation using RL. Later, in \cref{sec:modality_analysis}, we also see that this policy performs much worse on image and language goals. This shows that policy is having difficulty improving on these sub-tasks, causing the average performance (across all sub-tasks to be low). This trend of poor instance-specific image-goal performance is also evident when comparing image-goal policies trained using CLIP features vs. CroCo-v2 image features~\cite{croco_v2} (refer~\cref{sec:modality_analysis} for additional analysis).

\begin{table}
    \centering
    \resizebox{1\linewidth}{!}{
        \setlength\tabcolsep{2pt}
        \begin{tabular}{@{}ll
        *{2}{>{\centering\arraybackslash}p{1.25cm}}
        *{1}{>{\centering\arraybackslash}p{0.25cm}}
        *{2}{>{\centering\arraybackslash}p{1.75cm}}
        *{1}{>{\centering\arraybackslash}p{0.25cm}}
        *{2}{>{\centering\arraybackslash}p{1.25cm}}
        *{1}{>{\centering\arraybackslash}p{0.25cm}}@{}}
            \toprule
            & & \multicolumn{2}{c}{\textsc{Val Seen}} & & \multicolumn{2}{c}{\textsc{Val Seen Synonyms}} & & \multicolumn{2}{c}{\textsc{Val Unseen}} \\
            \cmidrule{3-4} \cmidrule{6-7} \cmidrule{9-10}
            & Method & SR $(\mathbf{\uparrow})$ & SPL $(\mathbf{\uparrow})$
                & & SR $(\mathbf{\uparrow})$ & SPL $(\mathbf{\uparrow})$ & & SR $(\mathbf{\uparrow})$ & SPL $(\mathbf{\uparrow})$ \\
            \midrule
            \\[-10pt]
            %& \rownumber & Skill Chain & $13.20$ & $-$ & & $16.80$ & $-$ & & $12.65$ & $-$ \\
            & \textcolor{gray}{GOAT-GTSem} \cite{chang2023goat} & $\color{gray} 56.7$ & $\color{gray} 40.3$ & & $\color{gray} 58.4$ & $\color{gray} 43.5$ & & $\color{gray} 54.3$ & $\color{gray} 41.0$ \\
            \midrule
            
            & Modular GOAT \cite{chang2023goat} & $26.3$ & $\textbf{17.5}$ & & $33.8$ & $\textbf{24.4}$ & & $24.9$ & $\textbf{17.2}$ \\
            & Modular CLIP on Wheels \cite{gadre2022cows} & $14.8$ & $8.71$ & & $18.5$ & $11.5$ & & $16.1$ & $10.4$ \\
            % & \rownumber & ConceptFusion & $13.9$ & $-$ & & $-$ & $-$ & & $-$ & $-$ \\
            \midrule
            % & \rownumber & Skill Chain & $14.63$ & $-$ & & $13.73$ & $-$ & & $11.03$ & $-$ \\
            & SenseAct-NN Skill Chain & $\textbf{29.2}$ & $12.8$ & & $\textbf{38.2}$ & $15.2$ & & $\textbf{29.5}$ & $11.3$ \\ % Composite SPL
            & SenseAct-NN Monolithic & $16.8$ & $9.4$ & & $18.5$ & $10.1$ & & $12.3$ & $6.8$ \\
            % & IL Monolithic & $7.5$ & $4.8$ & & $7.2$ & $3.9$ & & $6.7$ & $3.1$ \\
            \bottomrule
            \end{tabular}
    }
    \caption{\textbf{Results}. Comparison of end-to-end RL and modular methods on GOAT-Bench \textsc{HM3D} benchmark on 3 evaluation splits: 1) \textsc{Val Seen}, 2) \textsc{Val Seen Synonyms}, 3) \textsc{Val Unseen}.}
    \label{tab:mod_vs_end2end_results}
    \vspace{-2.0em}
\end{table}

\section{Analysis}

Here, we further analyze the performance of the best-performing modular method against SenseAct-NN methods.

\subsection{How do agents perform on each modality?}
\label{sec:modality_analysis}

\begin{figure}
    \centering
    \includegraphics[width=1\linewidth]{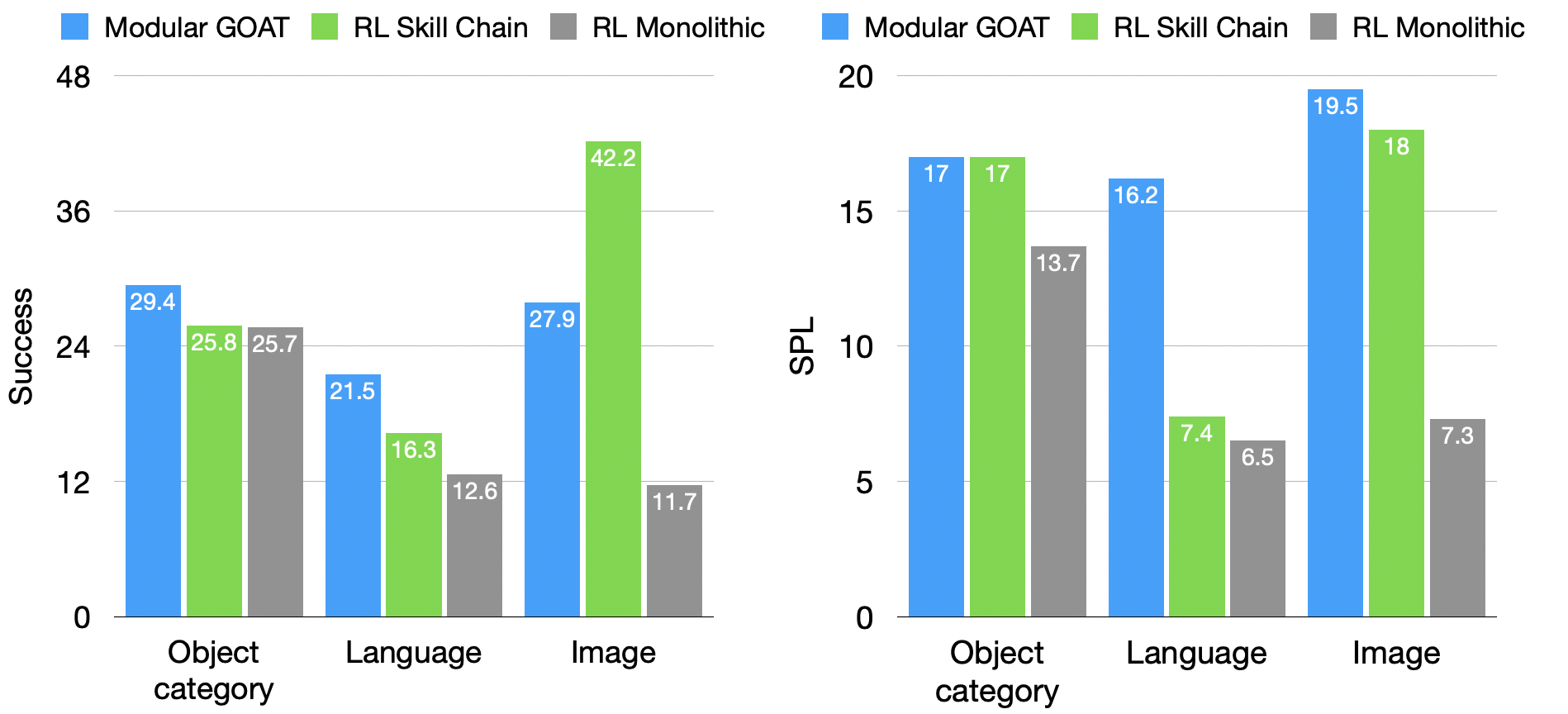}
    \caption{\textbf{Performance across types of modalities.} We breakdown the performance of all 3 baselines by modalities used subtask type: object category, language or image.}
    %We find that all baselines achieve the least success rate for navigating to language goals. Further, the RL skill chain baseline performs significantly better than other baselines on image goals. Finally, the end-to-end RL baseline finds navigating to object goals easier than navigating to other goals.}
    \label{fig:modality-wise_performance}
    \vspace{-2.0em}
\end{figure}

To understand how effective these agents are across the three modalities, we also plot modality-wise success rate and SPL numbers for the baselines on the \textsc{Val Seen} dataset in \cref{fig:modality-wise_performance}. For object goals, we observe that Modular GOAT \cite{chang2023goat} performs better on success rate than the SenseAct-NN Skill Chain and Monolithic baselines ($29.4\%$ vs $25.8\%$ and $25.7\%$). In terms of efficiency, we see that both Modular GOAT and SenseAct-NN Skill Chain perform equally well, and better than the SenseAct-NN Monolithic baseline. 
%Note that the RL skill chain baseline performs just as well on SPL on object and image goals – even without maintaining any scene memory.
%
For language goals, Modular GOAT performs better than SenseAct-NN Skill Chain on both – success (about $5\%$ better) and SPL (more than 2x better). This speaks to limitations of CLIP embeddings for capturing instance-specific features.

For image goals, we see that the CroCo-v2 Instance ImageNav policy used in the SenseAct-NN Skill Chain baseline outperforms Modular GOAT on success rate – by a huge margin of about $15\%$. On SPL, however, Modular GOAT does better because of its persistent memory. The SenseAct-NN Monolithic baseline, on the other hand, significantly underperforms on both success and SPL.
As shown in~\cref{fig:modality-wise_performance}, the ranks of the baselines using average performance is not indicative of performance across each modality. This is because, task-specific policies trained for SenseAct-NN Skill Chain baseline outperform other methods on \oviin task and is comparable on \ovon.

% \begin{table}
%     \centering
%     \resizebox{1\linewidth}{!}{
%         \setlength\tabcolsep{2pt}
%         \begin{tabular}{@{}cllrrcrrcrr@{}}
%             \toprule
%             & & & \multicolumn{2}{c}{\textsc{Object Category}} & & \multicolumn{2}{c}{\textsc{Language}} & & \multicolumn{2}{c}{\textsc{Image}} \\
%             \cmidrule{4-5} \cmidrule{7-8} \cmidrule{10-11}
%             & & Method & SR $(\mathbf{\uparrow})$ & SPL $(\mathbf{\uparrow})$
%                 & & SR $(\mathbf{\uparrow})$ & SPL $(\mathbf{\uparrow})$ & & SR $(\mathbf{\uparrow})$ & SPL $(\mathbf{\uparrow})$ \\
%             \midrule
%             \\[-10pt]
%             & \rownumber & GOAT & $29.4$ & $17.0$ & & $21.5$ & $16.2$ & & $27.9$ & $19.5$ \\
%             & \rownumber & Skill Chain & $25.8$ & $-$ & & $16.3$ & $-$ & & $42.2$ & $-$ \\
%             & \rownumber & End-to-End RL & $25.7$ & $13.7$ & & $12.6$ & $6.5$ & & $11.7$ & $7.3$ \\
%             \bottomrule
%             \end{tabular}
%     }
%     \vspace{5pt}
%     \caption{Comparison on success per modality on VAL SEEN split GOAT baselines.}
%     \label{tab:subtask_eval}
% \end{table}

% \vspace{-1.0em}
\subsection{How important is memory for efficient navigation?}

\begin{figure}
    \centering
    \includegraphics[width=1\linewidth]{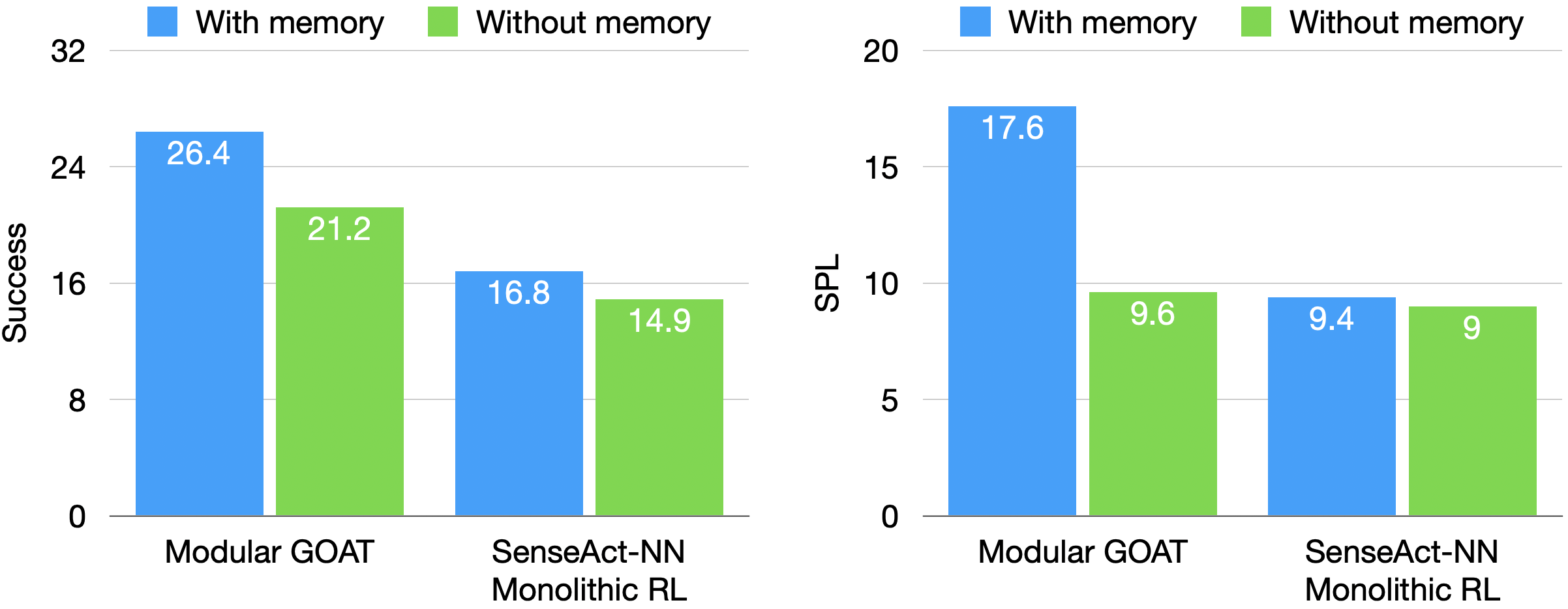}
    \caption{\textbf{Usefulness of memory}: We benchmark the drop in performance for when no memory is maintained across subtasks for modular GOAT~\cite{chang2023goat} and SenseAct-NN Monolithic RL baselines.}
    %- necessitating exploration from scratch post each sub-task. We find that the success path length (SPL) for GOAT~\cite{chang2023goat} baseline drastically increases when memory is erased between the sub-tasks - highlighting the importance of maintaining a memory.}
    \label{fig:memory_ablations}
    \vspace{-1.0em}
\end{figure}

Memory or the ability to remember previously seen objects or parts of the house can enable agents to be more efficient at navigation. For instance, an agent that has already seen the kitchen is expected to navigate directly to it (without exploring) when asked to find an oven. For methods using memory (\ie modular GOAT and monolithic policies), we evaluate the importance of memory towards success rate and efficiency by dropping the memory after each subtask. For Modular GOAT, we do this by building the map from scratch for each subtask, whereas for the SenseAct-NN Monolithic policy, we do this by dropping the hidden state between subtasks. This forces the policies to explore the environment from scratch for each subtask, and does not allow it to leverage past experience in the scene. As shown in \cref{fig:memory_ablations}, this results in a significant drop in SPL for Modular GOAT – by a factor of approximately 2x – from $17.6$ to $9.4$. The success rate also reduces by around $5\%$ (from $26.4$ to $21.2$). As highlighted in~\cite{chang2023goat}, success rate drops because a prebuilt scene memory leads to improved instance-to-goal matching. 
%This is because it is easier to find the goal instance using maximum match (from memory) as opposed to matching while the agent is exploring the scene using a predetermined threshold. If the threshold is too low, the agent might navigate to false positives, if it is too high, it might miss out on true positives.
%\simpletodo{@Mukul: Can you clarify this?}

\noindent The SenseAct-NN Monolithic policy, on the other hand, sees only a minor drop in SPL (from $9.4$ to $9.0$) and success (from $16.8$ to $14.9$) when memory is dropped. This suggests an inability (or lack of expressiveness) of the policy's hidden state to capture useful information about the explored scene. During evaluation, we often find the agent continuing to explore the scene when asked to navigate to an object it has seen previously (see \cref{sec:analysis_monolithic} for qualitative visualizations).
%

% \begin{table}
%     \centering
%     \resizebox{1\linewidth}{!}{
%         \setlength\tabcolsep{2pt}
%         \begin{tabular}{@{}llcccccccc@{}}
%             \toprule
%             & & \multicolumn{2}{c}{\textsc{Val Seen}} & & \multicolumn{2}{c}{\textsc{Val Unseen Easy}} & & \multicolumn{2}{c}{\textsc{Val Unseen Hard}} \\
%             \cmidrule{3-4} \cmidrule{6-7} \cmidrule{9-10}
%             & Method & SR $(\mathbf{\uparrow})$ & SPL $(\mathbf{\uparrow})$
%                 & & SR $(\mathbf{\uparrow})$ & SPL $(\mathbf{\uparrow})$ & & SR $(\mathbf{\uparrow})$ & SPL $(\mathbf{\uparrow})$ \\
%             \midrule
%             \\[-10pt]
%             & GOAT & $26.4$ & $17.6$ & & $-$ & $-$ & & $24.9$ & $17.2$ \\
%             % & \rownumber & GOAT no instance memory & $-$ & $-$ & & $-$ & $-$ & & $-$ & $-$ \\
%             & GOAT no memory & $21.2$ & $9.6$ & & $-$ & $-$ & & $-$ & $-$ \\
%             \midrule
%             \\[-10pt]
%             & End-to-End RL & $16.8$ & $9.4$ & & $18.5$ & $10.1$ & & $12.3$ & $6.8$ \\
%             % & \rownumber & GOAT no instance memory & $-$ & $-$ & & $-$ & $-$ & & $-$ & $-$ \\
%             & E2E RL no memory & $14.9$ & $9.0$ & & $17.0$ & $9.1$ & & $10.7$ & $6.7$ \\
%             \bottomrule
%             \end{tabular}
%     }
%     \vspace{5pt}
%     \caption{Ablations of memory for all baselines.}
%     \label{tab:modular-ablations}
% \end{table}

\subsection{Does success and efficiency improve over time?}

\begin{figure}
    \centering
    \includegraphics[width=1\linewidth]{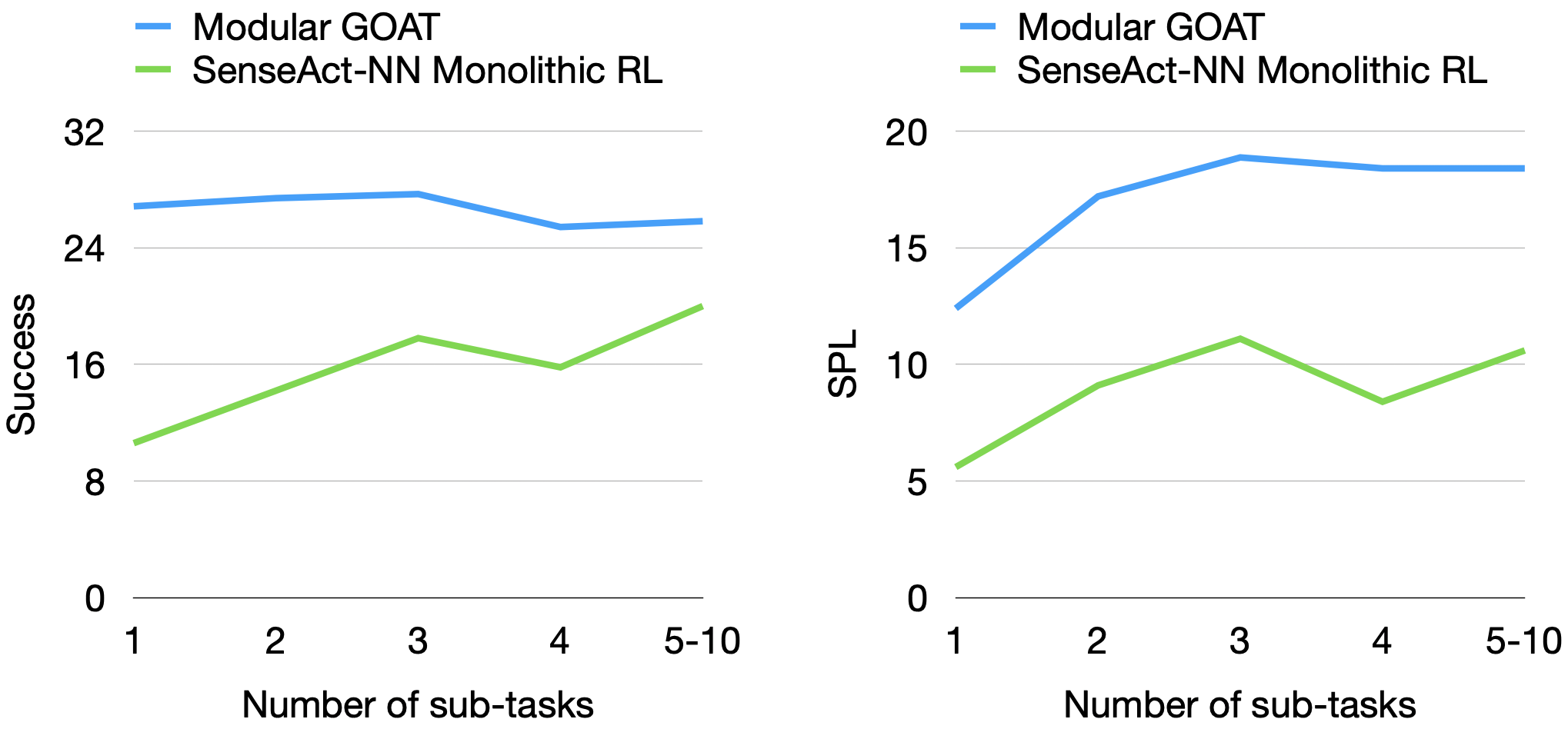}
    \caption{\textbf{Average performance over time in a GOAT episode.} We plot the success rate and SPL of memory based baselines against the number of subtasks completed in a GOAT episode.}
    %We find that goals appearing later in the episode result in better success with End-to-end RL and better SPL with GOAT~\cite{chang2023goat}, when compared to goals appearing earlier in the episode.}
    \label{fig:performance_over_time}
    \vspace{-1.8em}
\end{figure}

As agents perform more subtasks in the same environment, it is reasonable to expect them to get better over time. Efficient agents will ideally keep track of already seen objects and areas of the house and will have an internal model of paths to follow to reach already seen goals. To evaluate this, we plot average success and SPL over number of subtasks in an episode for these methods in \cref{fig:performance_over_time}.

We observe that for Modular GOAT, the success rate does not improve over subtasks, whereas the SPL does see gains over the first three subtasks (from $12.4$ to $18.7$) before it roughly saturates (at $18.4$).
For the SenseAct-NN Monolithic policy, both SPL and success rate does see gains over time, from $5.6$ to $10.6$ on SPL and $10.6\%$ to $20.0\%$ on success. 
These results highlight the importance of having effective memory representations (implicit or explicit) to perform efficiently on the GOAT task.

\subsection{How robust are these methods to noise in goal specifications?}
\label{sec:exp_noise_analysis}

\begin{figure}
    % \centering
    % \includegraphics[width=1\linewidth]{fig/analysis/noisy_evals.png}
    % \caption{\textbf{Evaluating robustness to noise.} We benchmark the affect of noise in goal specification on the performance of different baselines. We find that GOAT~\cite{chang2023goat} is significantly more sensitive to noise in goals when compared to other baselines, resulting in close to 50\% drop in success rate and SPL.}
    % \label{fig:noisy_evals}
    
    \centering
    \includegraphics[width=1\linewidth]{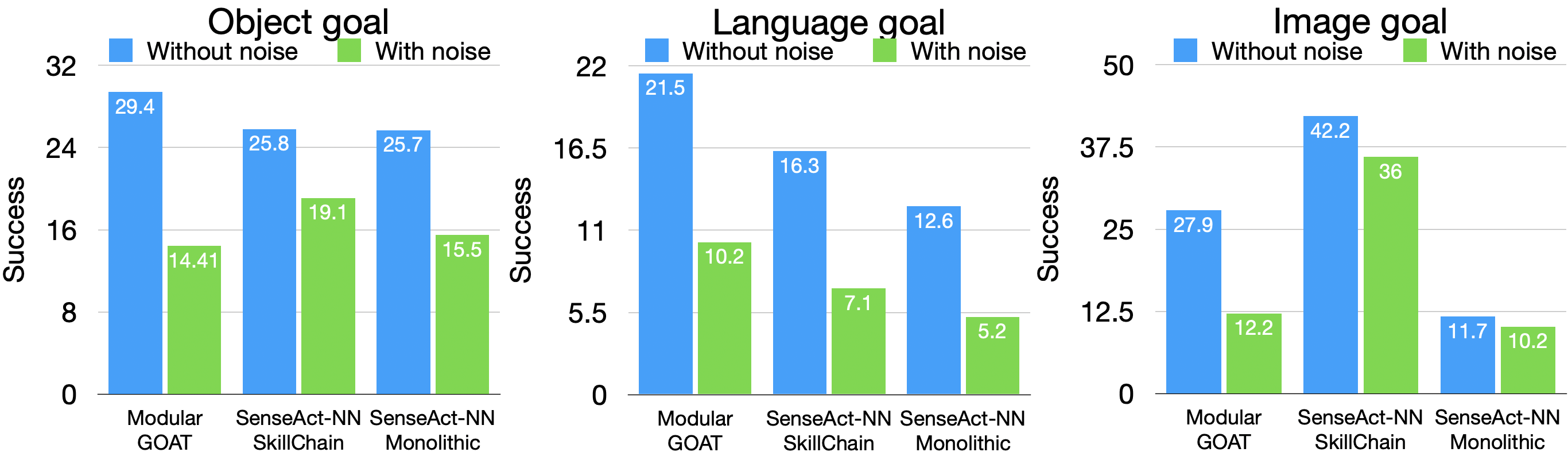}
    \caption{\textbf{Robustness to noise.} We breakdown the effect of noise on performance of different baselines by goal modality.}
    \label{fig:noisy_evals_per_modality}
    \vspace{-2.0em}
\end{figure}

Goal specifications in real-world scenarios can often contain a lot of noise. Images of goal object can be noisy (for example in low-lit scenes), users might use uncommon synonyms to describe object categories, or they might phrase descriptions of instances differently. To simulate this type of noise in goal specifications, we perturb the goal inputs of the three modalities in the following ways. We add gaussian noise ($\mu=0$, $\sigma=x \sim \mathcal{U}(0.1, 2.0)$)  to goal images, replace object category names with corresponding synonyms and paraphrase language descriptions of instances (using ChatGPT). We evaluate the baselines on the \textsc{Val Seen} split of the dataset and report their performance and robustness to noise – across three modalities – in \cref{fig:noisy_evals_per_modality}.

For object goals, we observe that Modular GOAT faces the biggest drop in performance when the object categories are replaced with synonyms. We find that this is because the object detector (DETIC \cite{zhou2022detecting} here) performs poorly on detecting these relatively uncommon synonyms. On the other hand, the skill chain and monolithic baselines don't suffer as much because they use CLIP goal embeddings (which capture these concepts better). %, \red{refer~\cref{fig:noise_examples} for examples.}
For language goals, all three baselines suffer a reasonable drop in success rate. This can be attributed to the lack of instance-specific expressiveness of CLIP embeddings that are used as goal embeddings for the RL baselines and for goal matching in Modular GOAT. %, \red{\cref{fig:noise_examples} shows an example of such case.}
For image goals, the SenseAct-NN methods suffer very little with gaussian noise. This speaks to the robustness of the visual features from cross-view consistent representations from the CroCo-v2 encoder~\cite{croco_v2} used for the SenseAct-NN Skill Chain baseline and CLIP used for the monolithic policy.
Overall, we observe that SenseAct-NN Skill Chain baseline is the most robust to noise (with a $25\%$ average drop in success), whereas GOAT is the least robust ($53\%$ drop).

% \begin{table}
%     \centering
%     \resizebox{1\linewidth}{!}{
%         \setlength\tabcolsep{2pt}
%         \begin{tabular}{@{}cllrrcrrcrr@{}}
%             \toprule
%             & & & \multicolumn{2}{c}{\textsc{Val Seen}} & & \multicolumn{2}{c}{\textsc{Val Unseen Easy}} & & \multicolumn{2}{c}{\textsc{Val Unseen Hard}} \\
%             \cmidrule{4-5} \cmidrule{7-8} \cmidrule{10-11}
%             & & Method & SR $(\mathbf{\uparrow})$ & SPL $(\mathbf{\uparrow})$
%                 & & SR $(\mathbf{\uparrow})$ & SPL $(\mathbf{\uparrow})$ & & SR $(\mathbf{\uparrow})$ & SPL $(\mathbf{\uparrow})$ \\
%             \midrule
%             \\[-10pt]
%             & \rownumber & GOAT & $12.6$ & $8.3$ & & $-$ & $-$ & & $-$ & $-$ \\
%             & \rownumber & Skill Chain & $-$ & $-$ & & $-$ & $-$ & & $-$ & $-$ \\
%             & \rownumber & End-to-End RL & $10.1$ & $5.9$ & & $-$ & $-$ & & $-$ & $-$ \\
%             \bottomrule
%             \end{tabular}
%     }
%     \vspace{5pt}
%     \caption{Comparison on robustness to noise in goal specification for GOAT baselines. }
%     \label{tab:noisy_evals}
% \end{table}

\vspace{-0.8em}
\section{Conclusion}

\looseness=-1 In this work, we propose GOAT-Bench, a novel reproducible benchmark for building and evaluating multi-modal lifelong navigation systems.
We believe, this benchmark is a step towards building general purpose navigation agents that can handle multi-modal goals (\eg image of an object, language description, and object categories) and leverage past experiences in the environment to perform the task efficiently.
On GOAT-Bench, we benchmark two classes of methods, modular and end-to-end trained methods with and without memory representations.
We find methods with effective memory representations perform well on GOAT task and achieve higher efficiency compared to methods without memory.
In addition, we present a comprehensive analysis of dependency of these methods on memory, performance across modalities, and robustness to noise in goals specified via different modalities.

\xhdr{Acknowledgements}.
The Georgia Tech effort was supported in part by NSF, ONR YIP, and ARO PECASE. The views and conclusions contained herein are those of the authors and should not be interpreted as necessarily representing the official policies or endorsements, either expressed or implied, of the U.S. Government, or any sponsor.

{\small
\bibliographystyle{ieeetr}
\setlength{\bibsep}{0pt}
\bibliography{main}
}

\clearpage
\newpage
\clearpage
\appendix
% \section {OVON Categories}
\label{sec:ovon_categories}
% \subsection{OVON Train Categories}
\begin{table*} [htp]
    \centering
    \begin{subtable}[c]{.8\textwidth}
    \centering
    \begin{tabular}{@{\extracolsep{\fill}} p{\linewidth} @{}}
\toprule
\textit{\scriptsize air conditioner,
amplifier,
antique clock,
antique telephone,
aquarium,
arcade game,
archway,
artwork,
baby changing station,
backrest,
backsplash,
bag,
balcony railing,
balustrade,
banner,
bar,
bar cabinet,
barbecue,
basket,
bath towel,
bathrobe,
bathroom accessory,
bathroom cabinet,
bathroom shelf,
bathroom towel,
bathtub,
bathtub platform,
bed,
bed comforter,
bed curtain,
bench,
binder,
blanket,
board,
board game,
book cabinet,
book rack,
bottle,
box,
brochure,
bucket,
bulletin board,
bunk bed,
cabinet,
cabinet clutter,
cabinet table,
candle,
candle holder,
canvas,
cardboard,
cardboard box,
cart,
case,
casket,
chair,
chimney,
cleaning clutter,
clock,
closet,
closet shelf,
closet shelving,
cloth,
clothes,
clothes bag,
clothes dryer,
clothing stand,
clutter,
coat,
container,
cooker,
copier machine,
cosmetics,
couch,
counter,
countertop,
cradle,
crate,
crib,
curtain valence,
cutting board,
decorative plate,
decorative quilt,
desk,
desk cabinet,
desk lamp,
dinner table,
dish cabinet,
dish rack,
dishrag,
display cabinet,
display table,
double armchair,
drawer,
drawer sink table,
dressing table,
duct,
easy chair,
electrical controller,
elephant sculpture,
elevator,
exercise ladder,
exhibition panel,
fence,
figure,
fire extinguisher,
firewood,
fish tank,
flag,
folding stand,
folding table,
foosball game table,
foot spa,
fuse box,
gas furnace,
gate,
globe,
grass,
grill,
guitar frame,
gym equipment,
hammock,
handbag,
handle,
hat,
headboard,
heater,
high shelf,
hook,
hose,
hunting trophy,
hutch,
icebox,
iron board,
jacket,
jacuzzi,
jar,
jewelry box,
kitchen appliance,
kitchen cabinet lower,
kitchen countertop items,
kitchen extractor,
kitchen lower cabinet,
kitchen shelf,
kitchen sink,
kitchen table,
kitchen top,
ladder,
lamp stand,
lampshade,
laptop,
laundry,
laundry basket,
light switch,
locker,
lounge chair,
magazine,
mantle,
massage bed,
medical lamp,
mixer,
monitor,
motorcycle,
newspaper,
note,
office chair,
office table,
ornament,
oven,
oven and stove,
painting frame,
pantry,
paper,
paper storage,
paper towel,
patio chair,
photo stand,
pile of magazines,
pillar,
plate,
plush toy,
pool,
pool table,
pot,
pouffe,
power breaker box,
purse,
rack,
radio,
range hood,
record player,
relief,
robe,
rocking chair,
round chair,
sauna oven,
scarf,
schedule,
screen,
seat,
sewing machine,
shade,
sheet,
shelf cubby,
shelving,
shower,
shower bar,
shower cabin,
shower rail,
shower soap shelf,
shower stall,
shower tap,
shower-bath cabinet,
sideboard,
sign,
sink,
sink cabinet,
sink table,
sled,
sleeping bag,
sofa chair,
sofa seat,
sofa set,
soft chair,
solarium,
spa bench,
spice rack,
stack of papers,
stack of stuff,
stage,
staircase trim,
stand,
stereo set,
stone support structure,
stool,
storage,
storage cabinet,
storage shelving,
storage space,
stove,
stovetop,
sunbed,
support beam,
swivel chair,
table,
table stand,
tank,
telephone,
telescope,
tent,
tile,
toilet,
tool,
towel,
toy,
trampoline,
trashcan,
tray,
treadmill,
tv,
umbrella,
urinal,
vacuum cleaner,
violin case,
wardrobe,
washbasin,
washbasin counter,
washer-dryer,
washing machine,
water dispenser,
water fountain,
water tank,
whiteboard,
window,
window shade,
window shutter,
wine cabinet,
wood,
workstation,
worktop,
wreath}
\\
\bottomrule
\end{tabular}
\caption{Training Split}
\vspace{1em}
\label{subtab:ovon_train_categories}
\end{subtable}
    \begin{subtable}[c]{.8\textwidth}
    \centering
    \begin{tabular}{@{\extracolsep{\fill}} p{\linewidth} @{}}
\toprule
\textit{\scriptsize bag,
balustrade,
basket,
bath towel,
bathrobe,
bathroom cabinet,
bathroom towel,
bathtub,
bathtub platform,
bed,
bench,
blanket,
board,
box,
cabinet,
cabinet table,
cardboard box,
case,
chair,
closet,
closet shelf,
cloth,
clothes,
clothes dryer,
container,
cooker,
couch,
countertop,
crib,
desk,
display cabinet,
drawer,
fire extinguisher,
flag,
handbag,
heater,
high shelf,
iron board,
kitchen appliance,
kitchen countertop items,
kitchen lower cabinet,
kitchen shelf,
mantle,
monitor,
note,
office chair,
oven,
oven and stove,
pillar,
plush toy,
rack,
shower,
shower cabin,
shower rail,
shower soap shelf,
sideboard,
sink,
sink cabinet,
sofa chair,
sofa seat,
sofa set,
spa bench,
stool,
storage shelving,
stove,
support beam,
table,
toilet,
towel,
toy,
tv,
wardrobe,
washbasin,
washer-dryer,
washing machine,
window,
window shade,
window shutter,
worktop}
\\
\bottomrule
\end{tabular}
\caption{Val Seen Split}
\vspace{1em}
\label{subtab:ovon_val_seen_categories}
\end{subtable}
    % \subsection{OVON Val Seen Synonyms}
\begin{subtable}[c]{.8\textwidth}
    \centering
    \begin{tabular}{@{\extracolsep{\fill}} p{\linewidth} @{}}
\toprule
\textit{\scriptsize appliance,
armchair,
backpack,
bar chair,
bath cabinet,
bath sink,
bathroom counter,
bed sheet,
bed table,
bedside lamp,
bicycle,
bookshelf,
chest drawer,
chest of drawers,
clothes rack,
coffee table,
computer,
computer chair,
computer desk,
curtain,
curtain rod,
cushion,
desk chair,
dining chair,
dining table,
electric box,
exercise equipment,
file cabinet,
fireplace,
folding chair,
furnace,
ironing board,
kitchen cabinet,
kitchen counter,
kitchen island,
lamp,
lamp table,
railing,
refrigerator cabinet,
shelf,
shower curtain,
shower tub,
stairs railing,
table lamp,
tablecloth,
throw blanket,
tv stand,
water heater,
window curtain,
window frame}
\\
\bottomrule
\end{tabular}
\caption{Val Seen Synonyms Split}
\vspace{1em}
\label{subtab:ovon_val_seen_syn_categories}
\end{subtable}
    % \subsection{OVON Val Unseen}
\begin{subtable}[c]{.8\textwidth}
    \centering
    \begin{tabular}{@{\extracolsep{\fill}} p{\linewidth} @{}}
\toprule
\textit{\scriptsize bedframe,
blinds,
boiler,
book,
bowl of fruit,
calendar,
carpet,
christmas tree,
clothes hanger rod,
coffee machine,
decorative plant,
dishwasher,
dresser,
exercise bike,
flower vase,
flowerpot,
food,
footrest,
footstool,
freezer,
glass,
guitar,
handrail,
hanger,
hanging clothes,
island,
microwave,
mirror,
nightstand,
ottoman,
parapet,
photo,
photo mount,
piano,
picture,
pillow,
plant,
printer,
radiator,
refrigerator,
rug,
shower dial,
shower glass,
speaker,
stair,
staircase handrail,
statue,
vase,
window glass}
\\
\bottomrule
\end{tabular}
\caption{Val Unseen Split}

\label{subtab:ovon_val_unseen_categories}
\end{subtable}
    \caption{Full list of object categories for Open-Vocab ObjectNav dataset \cite{ovon}.}
    \label{tab:ovon_categories}
\end{table*}

\section{Open-Vocab Instance-ImageNav dataset}
\label{sec:iinav_dataset}
%% Example figures for OVIIN

% \begin{figure*}[!ht]   
%     \centering
%     \includegraphics[width=0.7\textwidth]{fig/supp/oviin_examples.pdf}
%     \caption{Example image goals from Open-Vocab Instance-ImageNav dataset.}
%      % \simpletodo{@Mukul: Make figure consistent with other figures.}
%     \label{fig:oviin_examples}
% \end{figure*}

For creating the Open-Vocab Instance ImageNav (OVIIN) dataset, we build upon the dataset generation pipeline presented in \cite{krantz2022instance}, while increasing the diversity of object categories and instances. We start by using object categories and instances from the Open-Vocabulary ObjectNav dataset \cite{ovon}, instead of using just the 6 canonical ObjectNav categories \cite{habitatchallenge2023}. Refer \cref{tab:ovon_categories} for a full list of categories. For each object instance, we then generate image goals by capturing images from a set of candidate viewpoints around the object. Next, we filter out invalid image goals based on the object's visibility – through frame and object coverage heuristics.
We use the same parameters for sampling and thresholds same as the Instance-ImageNav (IIN) dataset \cite{krantz2022instance}. \cref{fig:more_goat_examples} shows more examples of the diverse set of image goals from the resultant OVIIN dataset.

\section{LanguageNav Dataset}
\label{sec:lnav_dataset}

As outlined in \cref{sec:dataset}, we sample goal instances for the LanguageNav dataset from the OVON dataset \cite{ovon}. We start by generating viewpoints for these goal instances – for effectively procuring language description annotations for them. This involves
sampling candidate viewpoints within a radius $r\in [1.0, 1.5, 2.0]$ for every $10^\circ$ sections around the centroid of the object. Next, we collect $512 \times 512$ resolution images (using HelloRobot's Stretch embodiment) from these viewpoints and filter out images from which the target object is not sufficiently visible – by computing the object's frame coverage. 
Next, we feed the viewpoint image with the highest frame coverage into the BLIP-2 model \cite{li2023blip2} to obtain a detailed description of the target object, using the following prompt:

\texttt{Question: describe the <category>? Answer:}.

\noindent Next, we leverage the semantics of nearby objects to generate meaningful captions. We do so by using ground truth semantics and depth information from the simulator – by selecting objects at least $4.5m$ away (using average depth) and with total frame coverage exceeding $0.5\%$. For each object, we consolidate its bounding box coordinates, area coverage, object category, semantic index, and a boolean for whether the object is the target object or not.

\begin{table*}[htp]

\noindent\\\texttt{\scriptsize
Generate an informative and natural instruction to a robot agent based on the given information(a,b):
\\
a. Region Semantics: <bbox\_metadata>\\
b. Target Object Description: <obj\_category>:<BLIP-2 Caption>
\\
Based on the region semantics dictionary which contains information about 2d bounding boxes given in the form of (xmin,ymin,xmax,ymax) in a view of the target object where (0,0) is the top left corner of the frame and a description of the apperance of target object write an language instruction describing the location of the target object, <obj\_category>, spatially relative to other objects as references.
\\
There are some rules:
\\
Don't use any absolute values of the numbers, only use relative directions. Do not show bounding box coordinates in the output. Think of giving this as an instruction to a robot agent based on the given details. Add a prefix "Instruction: Find the .." or "Instruction: Go to .." to the generated instruction.
}
\caption{ChatGPT prompt used for LanguageNav dataset generation.}
\label{tab:gpt_prompt_template}
\end{table*}

\begin{figure*}[ht]   
    \centering
    \includegraphics[width=0.75\textwidth]{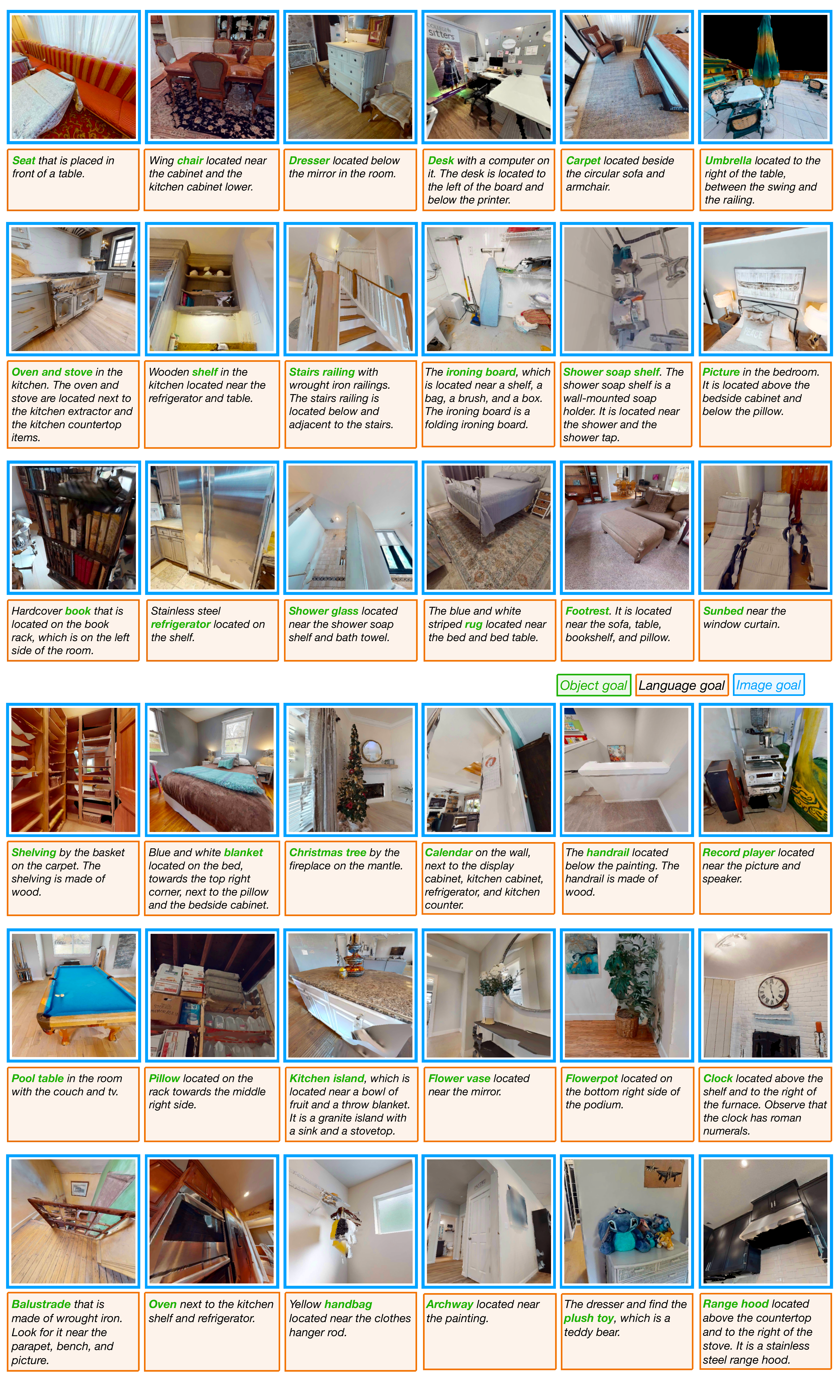}
    \caption{Additional multi-modal examples from the GOAT dataset.}
    \label{fig:more_goat_examples}
\end{figure*}

\noindent Finally, using the BLIP-2 generated caption and the bounding box metadata from the previous step, we create a prompt for ChatGPT, using the template shown in \cref{tab:gpt_prompt_template}. We use the \textit{GPT-3.5-Turbo} model and store the generated response as the language goal description for the target object. We present additional examples from the LanguageNav dataset in \cref{fig:more_goat_examples}.

\section{Baselines}

\begin{figure}
    \centering
    \includegraphics[width=1\linewidth]{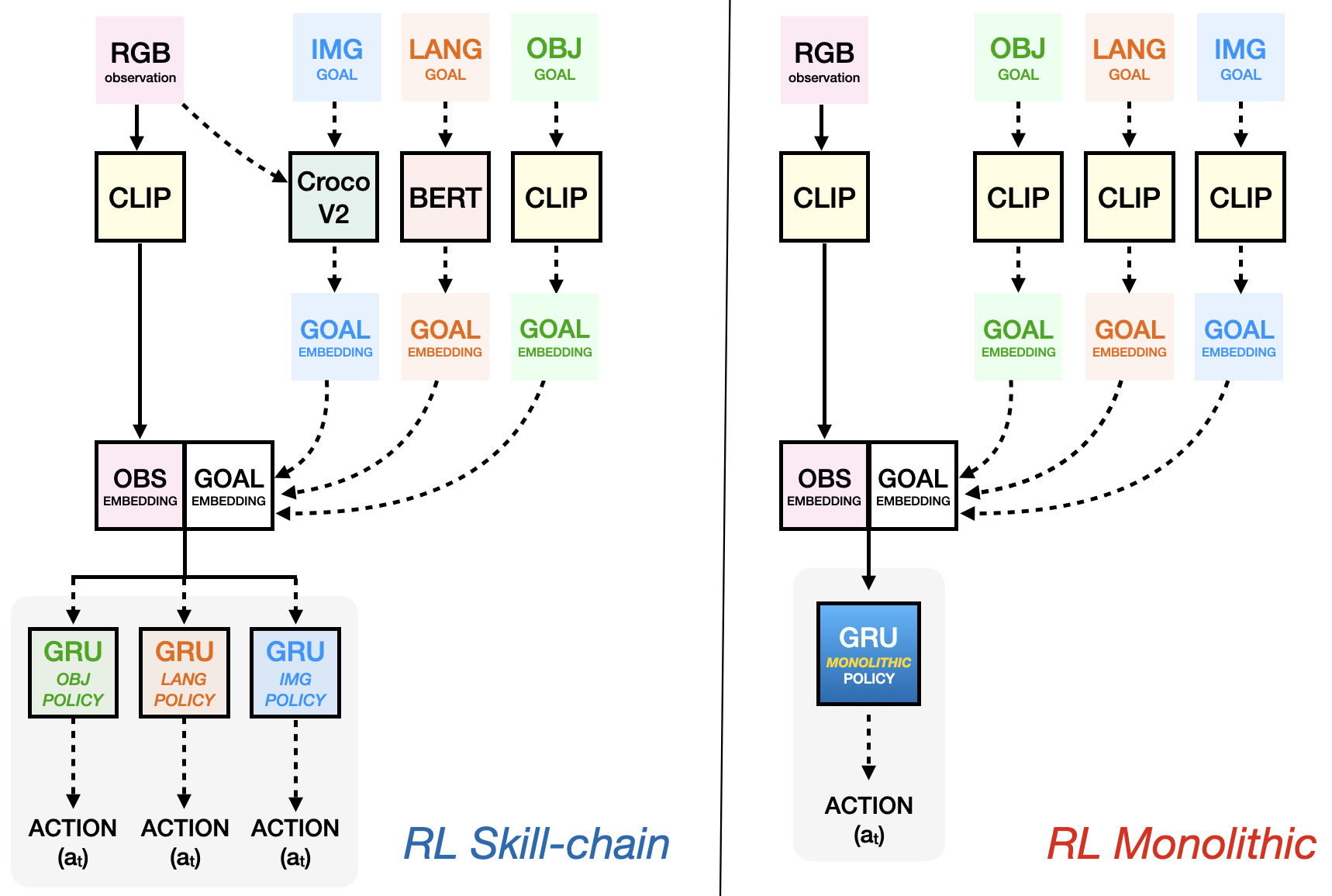}
    \caption{\textbf{Model architecture for SenseAct-NN Baselines}: We evaluate two types of SenseAct-NN RL policies: \textbf{Skill Chain} (left) and a \textbf{Monolithic Policy} architecture (right). Depending on the current goal specification (object category, language, or image), both methods first concatenate the goal +and current visual observation embeddings. The skill chain baseline then uses a policy pre-trained on the task corresponding to the current goal specification (e.g. ObjectNav policy for an object category goal). Once the current goal is reached, the hidden state is dropped and the agent switches to using a different policy depending on the subsequent goal. The monolithic policy, on the other hand, uses the same policy for all three types of goal specifications to predict the next action (while maintaining a hidden state memory).}
    \label{fig:baseline_architectures}
% \end{figure}

% \begin{figure}
    \centering
    
\end{figure}

We present architecture diagrams summarizing our end-to-end RL (skill chain and monolithic) and Modular GOAT baselines in \cref{fig:baseline_architectures} and \cref{fig:modular_goat_baseline_architecture}.

% \begin{figure}
%     \centering
%     \includegraphics[width=1\linewidth]{fig/supp/baseline_architectures.png}
%     \caption{}
%     \label{fig:baseline_architectures}
% \end{figure}

\subsection{SenseAct-NN Skill Chain}
\label{sec:skill_chain_hyperparam}

\begin{table}[ht]
    \centering
    \resizebox{1\linewidth}{!}{
     \setlength\tabcolsep{2pt}
        \begin{tabular}{@{}cllrrcrrcrr@{}}
            \toprule
            & & & \multicolumn{2}{c}{\textsc{Val Seen}} & & \multicolumn{2}{c}{\textsc{Val Unseen Easy}} & & \multicolumn{2}{c}{\textsc{Val Unseen Hard}} \\
            \cmidrule{4-5} \cmidrule{7-8} \cmidrule{10-11}
            & & Task & Success $(\mathbf{\uparrow})$ & SPL $(\mathbf{\uparrow})$
                & & Success $(\mathbf{\uparrow})$ & SPL $(\mathbf{\uparrow})$ & & Success $(\mathbf{\uparrow})$ & SPL $(\mathbf{\uparrow})$ \\
            \midrule
            \\[-10pt]
            & \rownumber & OVON & $32.5$ & $16.3$ & & $28.60$ & $14.0$ & & $15.70$ & $7.0$ \\
            % & \rownumber & LanguageNav & $14.63$ & $-$ & & $13.73$ & $-$ & & $11.03$ & $-$ \\
             & \rownumber & LanguageNav & $17.1$ & $7.8$ & & $16.9$ & $8.1$ & & $13.5$ & $6.1$ \\
            & \rownumber & OVIIN & $48.2$ & $22.3$ & & $45.8$ & $21.6$ & & $45.3$ & $22.3$ \\
            \bottomrule
            \end{tabular}
    }
    % \vspace{5pt}
    \caption{Performance of individual skills used for RL Skill Chain baseline on Open-Vocabulary ObjectNav \cite{ovon}, LanguageNav, and Open-Vocabulary Instance-ImageNav (OVIIN) val splits. We explain each of these policies in \cref{sec:skill_chain_hyperparam}.}
    \label{tab:individual_performance}
\end{table}

For our RL skill chain baseline, we train individual policies for \lnav, \objnav and \iinav tasks using VER~\cite{wijmans2022ver}. % The success distance is set to 0.25m.
Each policy takes in RGB observation size 224x224, which we obtain by resizing and center cropping the original $360 \times 640$ image input.
The agent also has access to GPS+Compass sensors, which provides location and orientation relative to the start of episode.
We embed the pose information to a 32-dimensional vector and concatenate with RGB observation embedding to form current state embedding.
Next, as shown in \Cref{fig:baseline_architectures}, we compute an N-dimensional goal embedding vector using a modality specific goal encoder (separate for each baseline), and concatenate it with state embedding to form a observation embedding $o_t$.
Finally, we feed the observation embedding with previous action into a $2-$layer 512-d GRU at each timestep to prediction a distribution over actions $a_t$.

\noindent \textbf{ObjectNav Policy}. For \objnav, we encode the object goal category to a target embedding $g_t$ (1024-d vector) using a frozen CLIP pretrained transformer based sentence encoder~\cite{radford2021learning}. We train the policy using RL with a navigation reward to minimize the distance-to-target. We train the policy till convergence (300 million steps in our case) using 4xA40 GPUs with 32 environments on each GPU. We present results of the model checkpoint with the best average performance across all $3$ evaluation splits of \ovon \textsc{HM3D} dataset.

\noindent \textbf{LanguageNav Policy}. For \lnav, we encode the language goal to a target embedding $g_t$ (768-d vector) using frozen BERT base uncased~\cite{bert} sentence encoder. Specifically, we use the output of the \texttt{[CLS]} token as language goal embedding $g_t$. 
%For these subtasks, the individual policies are trained for 300M steps across 24 environments and the best checkpoint on validation set is used.
Similar to the ObjectNav policy, we train the policy till convergence (300 million steps in our case) using 4xA40 GPUs with 32 environments on each GPU and choose the checkpoint with the best average performance across all $3$ evaluation splits.

\noindent \textbf{Instance ImageNav Policy}. To encode the instance image goal \iinav we use frozen CroCo-v2~\cite{croco_v2} image encoder, pre-trained on the Cross-view completion task on a dataset of 3D scanned scene images from Habitat~\cite{savva2019habitat} and real world images from ARKitScenes~\cite{arkitscenes}, MegaDepth~\cite{li2018megadepth}, 3DStreetView~\cite{3dstreetview} and IndoorVL~\cite{lee2021largescale} datasets.
Similar to a recent work~\cite{bono2023endtoend}, we also use adapter layers~\cite{chen2022adaptformer} with CroCo-v2 image encoder during training. 
%As the checkpoints for~\cite{bono2023endtoend} are not publicly available, we conduct RL fine-tuning to train our own policy. 
We use the publicly released pretrained CroCo-v2 ViT-Base Small-Decoder model from~\cite{croco_v2}.
Due to resource constraints, we resize and center crop the input image to $112\times 112$ during training.
%All other goal encoder parameters are maintained consistent with \cite{bono2023endtoend} and \cite{chen2022adaptformer}. 
We train the policy till convergence ($200$ million steps in our case) using 4xA40 GPUs with $16$ environments on each GPU. Similar to the ObjectNav and LanguageNav policies, we choose the checkpoint with the best average performance across all $3$ evaluation splits of \iinav dataset.
% The policy is trained on the Open-Vocab Instance-ImageNav dataset for 200 million steps across 16 environments. The best checkpoint on the validation set is used for skill-chaining.

\noindent The individual task performances are presented in \cref{tab:individual_performance}. We see that the InstanceImageNav (IIN) baseline performs the best across all the three tasks – indicating the efficacy of the cross-view consistent goal embeddings using the CroCo-V2 encoder \cite{croco_v2}. On the other hand, the LanguageNav task baseline struggles the most due to the inadequacy of CLIP features in capturing instance-specific information about objects.

\begin{figure*}   
    \centering
    \includegraphics[width=0.98\textwidth]{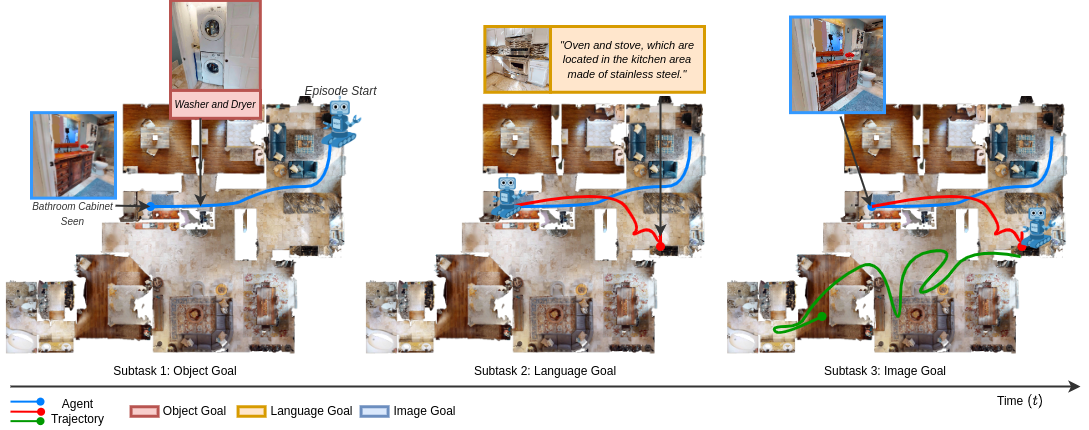}
    \caption{Qualitative example of the monolithic RL baseline agent not remembering objects or regions already seen in the environment. The agent starts by navigating to a washer dryer and sees the bathroom cabinet on its way. However, when tasked with navigating to the same bathroom cabinet in the third sub-task of this GOAT episode, the agent does not go back to the seen region of the house, but instead keeps exploring new regions.}
    \label{fig:goat_mono_eval}
\end{figure*}

\subsection{SenseAct-NN Monolithic Policy}
\label{sec:end_to_end_hyper}

\cref{fig:baseline_architectures} shows the architecture of the SenseAct-NN monolothic RL policy \ie a single end-to-end policy using multimodal goal encoder and implicit memory trained on the GOAT task.
The policy takes in RGB observation size $224 \times 224$, obtained by resizing and center-cropping the original input.
We encode the image $(i_t =$ CNN$(I_t))$ using a frozen CLIP~\cite{radford2021learning} ResNet50~\cite{he_cvpr16} encoder.
The agent also has access to GPS+Compass sensors, which provides location and orientation relative to the start of episode.
The GPS+Compass inputs, $P_t = (\Delta x, \Delta y, \Delta z)$, and $R_t = (\Delta \theta)$,
% between successive steps relative to start of the episode $(x_0, y_0, z_0, \theta_0)$.
%
are passed through fully-connected layers $p_t =$ FC$(P_t), r_t =$ FC$(R_t)$ to embed them to 32-d vectors.
Next, we compute a $1024$-d goal embedding vector $g^{k}_t$ using frozen CLIP image or sentence encoder based on the subtask $s_k$ goal modality (object, image, or language).
%modality specific goal encoder which is separate for each baseline, and concatenate it with state embedding to form a observation embedding $o_t$.
%
All these input features are concatenated to form an observation embedding, and fed into a 2-layer, 512-d GRU at every timestep to predict a distribution over actions $a_t$ - formally, given current observations $o_t = [i_t, p_t, r_t, g^{k}_t]$, $(h_t, a_t) =$ GRU$(o_t, h_{t-1}) $.
%
%Finally, we feed the observation embedding with previous action into a $2-$layer 512-d GRU at each timestep to prediction a distribution over actions $a_t$.
To leverage memory from the agent's past experiences in the scene, we carry forward hidden state of the policy from the last subtask $h^{(s_{t-1})}_{T}$ as initial hidden state for a new subtask $h^{(s_t)}_{0}$ in a single \goat episode.
% As described in~\cref{sec:e2e_baseline}, we use frozen CLIP image and sentence encoder for encoding goals (object, image, and language) of different modalities into a $1024-$d vector.
%
We train the policy till convergence using VER~\cite{wijmans2022ver} (for 500 million steps in our case) using 4xA40 GPUs with 32 environments on each GPU. We choose the checkpoint which has the best average performance across all $3$ evaluation splits of the GOAT \textsc{HM3D} dataset.

\subsection{Modular GOAT}

\begin{figure}
\centering
\includegraphics[width=1\linewidth]{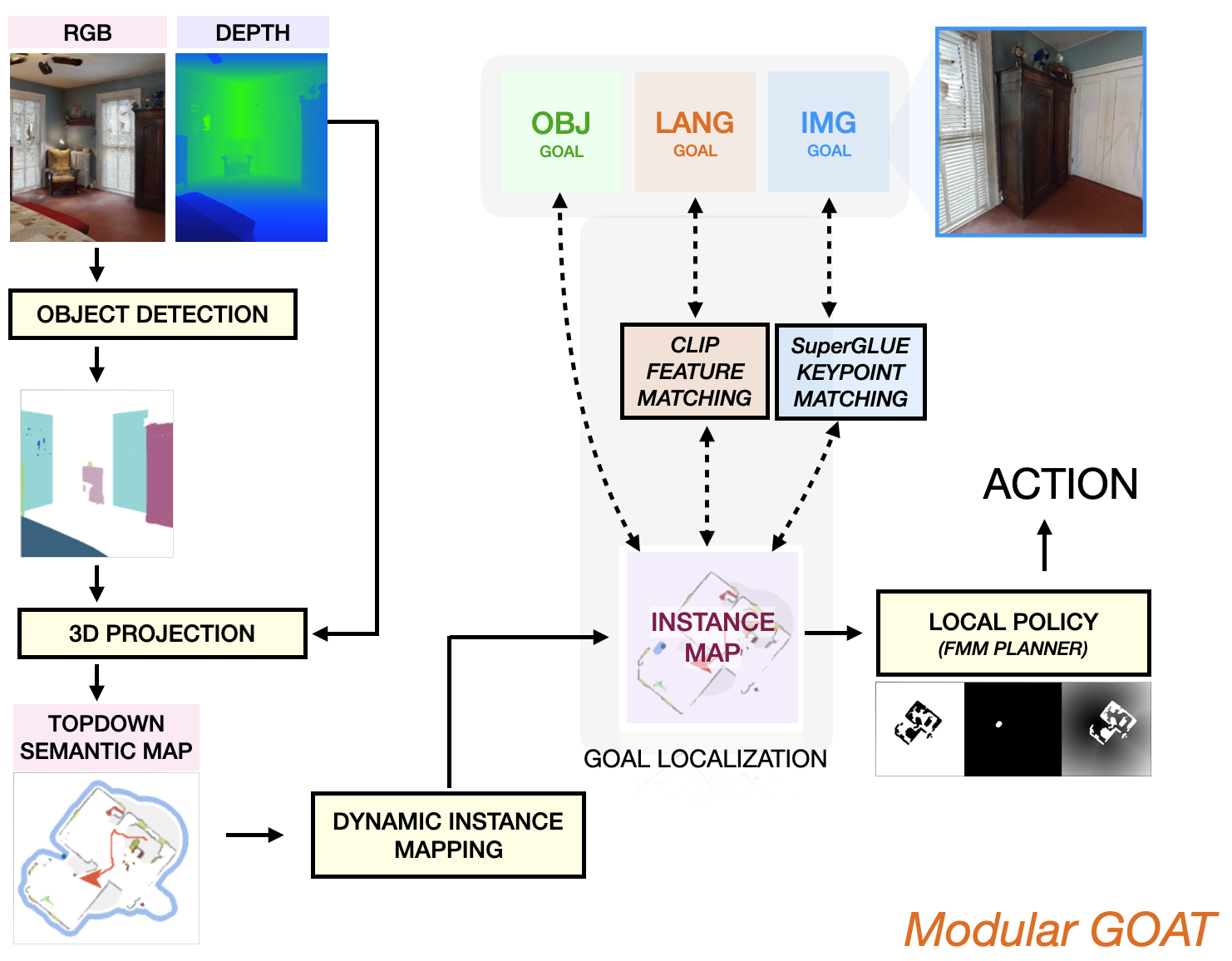}
    \caption{\textbf{Modular GOAT Baseline Architecture}: This baseline \cite{chang2023goat} maintains a semantic and instance-specific topdown map of the environment using a perception module combined with ground truth pose and depth information. This is then used to localize object, language, and image goals – by matching CLIP features or image keypoints. The agent explores the scene using frontier-exploration until a match is found. The goal is then passed to a local policy which predicts low-level actions to reach the goal.}
    \label{fig:modular_goat_baseline_architecture}
\end{figure}

We provide a visual overview of the Modular GOAT baseline, as proposed in \cite{chang2023goat}, in \cref{fig:modular_goat_baseline_architecture}. We direct the reader to prior work \cite{chang2023goat} for more information about the method.

% \section{Ablation SenseAct-NN Skill Chain}
% \label{sec:modular_rl}

% \section{Noise Analysis Details}
% \label{sec:noise_analysis}

% \begin{figure}[t]   
%     \centering
%     \includegraphics[width=0.75\linewidth]{example-image-a}
%     \caption{\textbf{Results.} Examples of noise added to object, image and language goals}
%     \label{fig:noise_examples}
% \end{figure}

\section{Qualitative Analysis of SenseAct-NN Monolithic Agent}
\label{sec:analysis_monolithic}

In this section, we discuss the inability of the SenseAct-NN monolithic RL policy in capturing past experiences in its implicit hidden state memory. Through \cref{fig:goat_mono_eval}, we share a qualitative example exhibiting this behaviour. The agent first looks for a washer dryer in the house (specified by object category). While navigating to the washer dryer, note that the agent also sees the bathroom cabinet shown in the figure. Ideally, we would expect the agent to keep track of this seen object when asked to navigate to it in the future. For the second sub-task, the agent successfully navigates to an oven and stove specified by language. However, for the third sub-task, when the agent it is tasked with navigating to the bathroom cabinet it saw before, the agent wanders around the house and does not make any attempt to visit the seen part of the house again. This highlights the lack of effectiveness of the GRU hidden state in the current implementation – in keeping track of seen objects and regions of the house.

\end{document}